\documentclass[11pt]{article}

\usepackage[final]{acl}

\usepackage{times}
\usepackage{latexsym}
\usepackage{amsmath,amssymb}

\usepackage[T1]{fontenc}

\usepackage[utf8]{inputenc}

\usepackage{microtype}

\usepackage{inconsolata}

\usepackage{graphicx}
\usepackage{diagbox}
\usepackage{booktabs}
\usepackage{multirow}
\usepackage[table]{xcolor}
\usepackage[most]{tcolorbox}
\usepackage{enumitem}

\usepackage[most]{tcolorbox}
\usepackage[dvipsnames, svgnames, x11names]{xcolor}

\definecolor{darkblue}{HTML}{1F33B4}
\definecolor{slightbule}{HTML}{77AFDF}
\tcbuselibrary{listings,skins,breakable}
\newtcblisting{templatebox}[1]{
    enhanced,
    colback=white,
    colframe=slightbule,
    colbacktitle=slightbule,
    coltitle=white,
    fonttitle=\bfseries,
    title=#1,
    halign=flush left,
    arc=2.5mm,
    boxrule=1pt,
    drop fuzzy shadow={gray!50!white},
    boxsep=0.8mm,
    left=1.5mm,
    right=1.5mm,
    top=1mm,
    bottom=2mm,
    listing only,
    listing options={
        breaklines=true,
        breakautoindent=false,
        breakindent=0pt,
        aboveskip=0pt,
        belowskip=0pt,
        columns=fullflexible,
        keepspaces=true,
        showstringspaces=false,
        tabsize=2
    }
}

\title{CoEvolve: Training LLM Agents via Agent-Data Mutual Evolution}

\author{
  Shidong Yang\textsuperscript{*},
  Ziyu Ma\textsuperscript{*},
  Tongwen Huang\textsuperscript{*},
  \textbf{Yiming Hu},
  \textbf{Yong Wang\textsuperscript{\textdagger}},
  \textbf{Xiangxiang Chu} \\
  AMAP, Alibaba Group \\
\url{https://github.com/AMAP-ML/CoEvolve}
}

\begin{document}
\maketitle
\begingroup
\renewcommand{\thefootnote}{\fnsymbol{footnote}}
  \footnotetext[1]{Equal contribution.}
  \footnotetext[2]{Project lead and corresponding author.}
\endgroup
\setcounter{footnote}{0}
\begin{abstract}

Reinforcement learning for LLM agents is typically conducted on a static data distribution, which fails to adapt to the agent's evolving behavior and leads to poor coverage of complex environment interactions. To address these challenges, we propose CoEvolve, an agent-data mutual evolution framework that enables LLM agents to improve through closed-loop, interaction-driven training. Specifically, CoEvolve extracts feedback signals such as forgetting and uncertainty from rollout trajectories to identify failure-prone interaction patterns, and utilizes them to guide LLM-based task synthesis. The synthesized tasks are validated through environment interaction and utilized to update the data distribution, enabling joint adaptation of the agent and its data. Extensive experiments on AppWorld and BFCL across Qwen2.5-7B, Qwen3-4B, and Qwen3-30B-A3B demonstrate consistent and significant improvements over strong base models, yielding absolute gains of 19.43\%, 15.58\%, and 18.14\%, respectively.
\end{abstract}

\section{Introduction}

The rapid advancement of large language models (LLMs) \cite{liu2024deepseek,yang2025qwen3, gou2025empirical} has driven the development of LLM-based agents, which have been widely applied to scenarios such as web information retrieval, software engineering, web navigation, and personal assistance \cite{jin2024llms,ding2025toolcoder, trivedi2024appworld, Ma_Gou_Hu_Wang_Zhuang_Cai_2026,ma2024drvideo}. Reinforcement learning (RL) \cite{guo2025deepseek,sun2024llm, ji2025tree,chu2025gpg} has emerged as the dominant approach for training these agents with complex interactive capabilities, offering a general solution for acquiring adaptive behaviors in open-ended environments.

\begin{figure}[t]
  \includegraphics[width=\columnwidth]{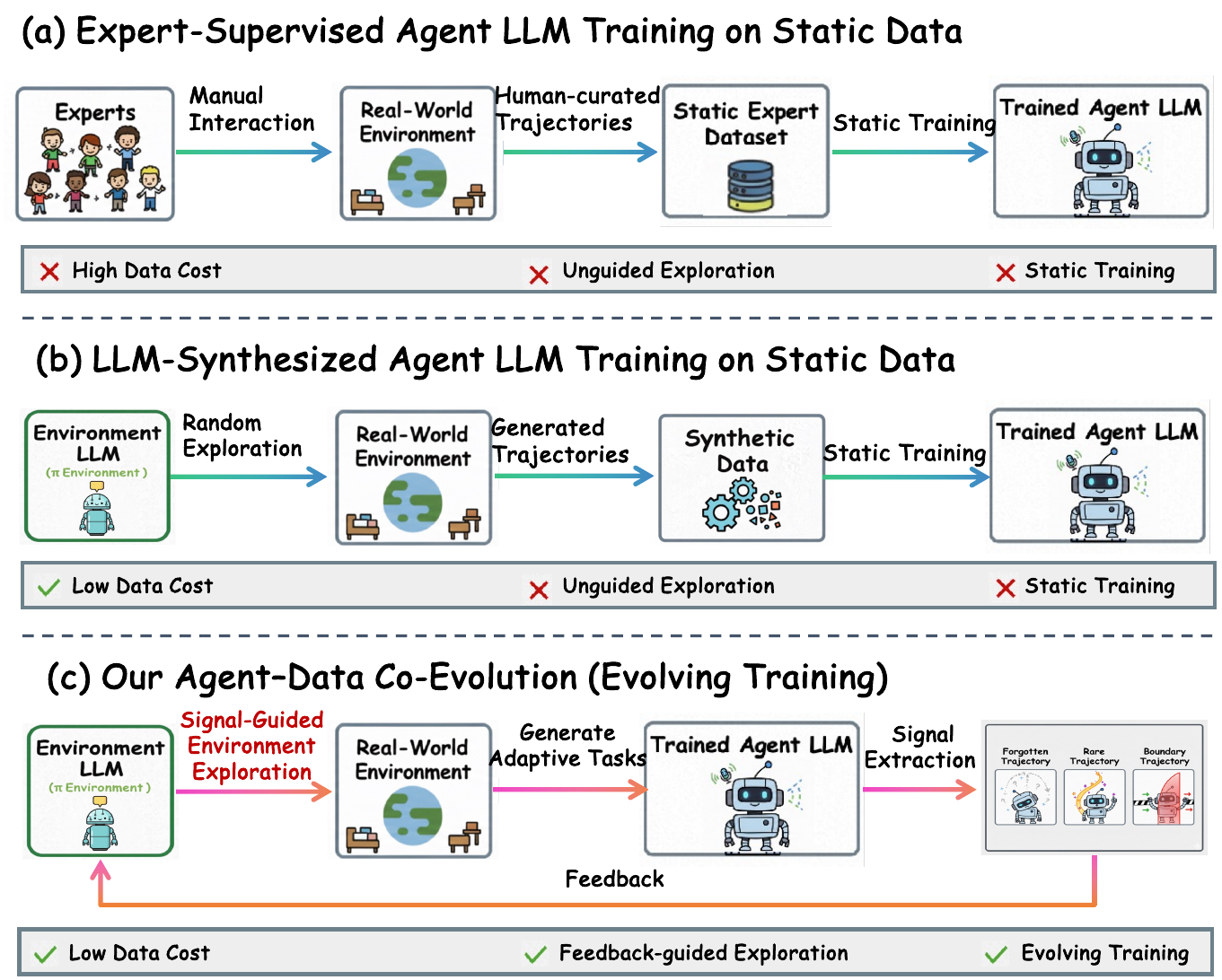}
  \caption{(a) Expert-Supervised. Agents learn from human-collected expert trajectories, incurring high data collection costs and limited generalization. (b) Static Synthetic. LLMs generate synthetic data in an offline and open-loop manner, yielding a static and non-adaptive training set. (c) Agent-Data Co-Evolution. Agents learn from tasks that evolve through feedback-driven interaction, enabling adaptive training without human supervision.}
  \label{fig:comparative_methods}
\end{figure}

However, current agent RL training methods \cite{li2025deepagent, mai2025agent, lin2025comprehensive} heavily rely on human-written demonstrations, where experts manually interact with the environment to construct trajectory datasets. These curated trajectories are then used to train the agent’s policy, as illustrated in Fig.~\ref{fig:comparative_methods}(a). While effective on simple tasks, this reliance on manually curated data introduces several critical limitations: (1) Collecting interaction data in real environment is prohibitively expensive, with a single trajectory often requiring several minutes or more of human expert effort. Given the limited availability of expert time, broad exploration of the environment becomes difficult.
(2) More fundamentally, these expert demonstrations represent static snapshots of interaction patterns and fail to cover the long-tail variations found in real-world settings \cite{wang2025co}. As a result, agents trained on such data struggle to generalize beyond the observed distribution. For instance, a web navigation agent may fail entirely if a button label changes from ``Book Now'' to ``Reserve Now''~\cite{gur2023understanding}.

The challenge of insufficient and static data has led to significant interest in synthetic data generation \cite{zhai2025agentevolver,mai2025cues,ding2024, ye2024}. A typical pipeline, illustrated in Fig.~\ref{fig:comparative_methods}(b), prompts a large language model (LLM) with environment descriptions and task specifications to explore the environment. By leveraging its world knowledge and reasoning capabilities, the LLM generates synthetic trajectories that are subsequently used to train the agent. While synthetic data reduces reliance on human annotation, it is typically generated through random exploration guided solely by the LLM’s world knowledge, without any feedback from the agent’s actual performance or interaction signals. Therefore, the environment exploration remains shallow and incomplete, failing to sufficiently cover diverse environment configurations. Moreover, the generated data still constitutes a static corpus that cannot adapt to the agent’s evolving capabilities, leading to inefficient training that neither targets specific weaknesses nor supports continual improvement.

To address these issues, we propose CoEvolve, an agent-data mutual evolution framework in which the agent and its training distribution evolve jointly through interaction-driven feedback, as shown in Fig.~\ref{fig:comparative_methods}(c). Our core idea is to use feedback signals, such as forgetting signals, to identify failure-prone interaction patterns and guide LLM-based task discovery accordingly. Unlike previous methods that rely on static datasets, CoEvolve synthesizes new tasks targeting the agent’s current weaknesses, validates them in the environment, and integrates them into training without human supervision. This closed loop allows the agent to reshape its learning distribution (data evolving) while continually overcoming its limitations (agent evolving).

We evaluate CoEvolve on two representative benchmarks, AppWorld \cite{trivedi2024appworld} and BFCL \cite{patilberkeley}, using Qwen2.5-7B, Qwen3-4B, and Qwen3-30B-A3B as backbones \cite{qwen25,yang2025qwen3}. By continuously synthesizing new tasks from training-time feedback, CoEvolve improves average performance by 19.43\%, 15.58\%, and 18.14\%, respectively, demonstrating strong scalability and generalization across models and environments. Our contributions can be summarized as follows:
\begin{itemize}
[leftmargin=*,topsep=1pt,itemsep=0.4pt]
\item 
We propose CoEvolve, an agent-data mutual evolution framework that alternates between agent optimization and data distribution updates without any human supervision.
\item 
Unlike previous synthetic data generation based on unguided random exploration, we incorporate feedback signals (e.g., forgetting signals) into LLM-based environment exploration.
\item 
CoEvolve yields large gains over baseline models (e.g., Qwen3-4B) across interactive benchmarks (e.g., AppWorld), demonstrating its effectiveness in complex environments.
\end{itemize}

\section{Related Work}
\textbf{Large Language Model Agents.}
Recent work has shown that large language models (LLMs) can be instantiated as autonomous agents capable of long-horizon reasoning and action through iterative interaction with environments.
Early frameworks such as ReAct~\citep{yao2022react} and Reflexion~\citep{shinn2023reflexion} demonstrate that coupling reasoning, tool use, and feedback enables LLMs to solve complex multi-step tasks, while later systems further enhance planning and memory for more persistent behaviors~\citep{zhang2024knowagent}.
Despite these advances, most existing LLM agents are trained via imitation learning on static collections of expert trajectories~\citep{nakano2021,wang2023}, which fundamentally limits exploration and constrains learning to the coverage of pre-collected data~\citep{shinn2023reflexion}.
In contrast, our work departs from this static paradigm by enabling agents to learn in a dynamic, self-evolving training process without relying on fixed expert demonstrations.

\begin{figure*}[t]
  \includegraphics[width=\textwidth]{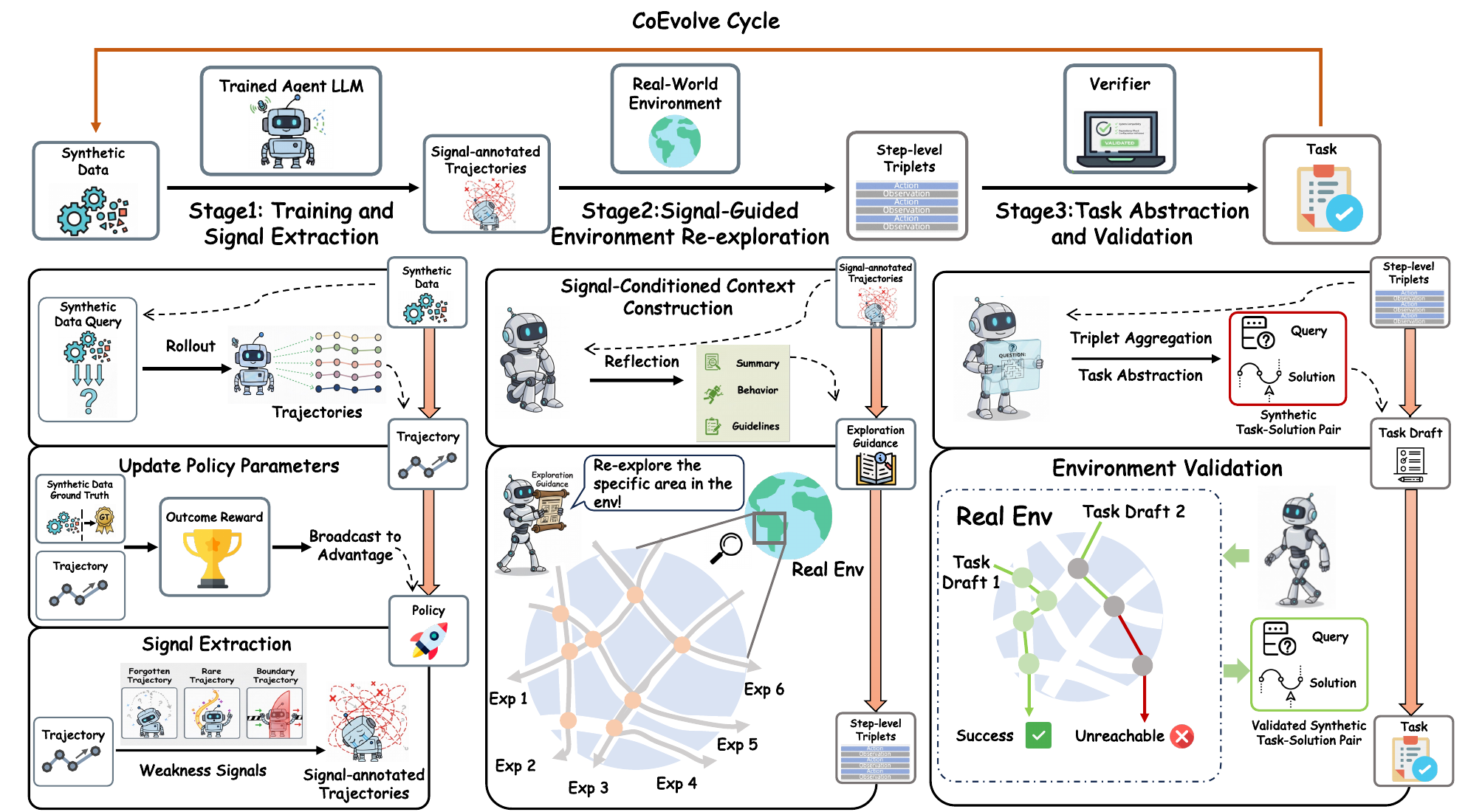}
\caption{
Overview of the \textbf{CoEvolve} framework. The agent is trained with GRPO, and feedback signals are extracted from rollout trajectories (Stage~1). These signals guide signal-conditioned re-exploration via an LLM (Stage~2) and are transformed into validated tasks to evolve the training set (Stage~3). This closed-loop process enables CoEvolve without human supervision.
}
\label{fig:co_evolve}
\end{figure*}

\noindent\textbf{Trajectory Synthesis for Agent Training.}
To reduce reliance on expert demonstrations, recent work explores synthetic trajectory generation for training LLM agents~\citep{yu2025demystifying}.
Most prior approaches generate trajectories in an \emph{offline} or weakly adaptive manner, including open-loop synthesis with reflection or correction~\citep{ye2024,ding2024,chen2025,chen2025stepwise}, as well as large-scale pipelines based on tutorials, scripted exploration, simulators, and self-training~\citep{pahuja2025,yuan2024,hoang2025,yuan2025,wang2025co,zhang2025,luo2025}.
Recent extensions introduce more autonomous exploration or structured curricula~\citep{wang2025uisimulator,zhao2025autoplay,zhang2025deepanalyze,liu2025limi,chen2025compo,sun2025earlyexperience}, yet trajectory generation remains largely \emph{open-loop}, loosely coupled to the agent’s evolving failure modes. In contrast, our method closes this loop by using environment feedback to synthesize trajectories on demand, enabling continuous adaptation of the training distribution.
Conceptually, CoEvolve also differs from recent self-improving or curriculum-style frameworks that refine trajectories for a fixed pool of queries or generate variants around seed tasks. Our feedback is used to drive the agent back into the interactive environment to discover new executable queries and states, so data evolution is not limited to rewriting or filtering an offline query set.

\section{Method}
\label{sec:method}

We propose CoEvolve, an agent-data co-evolution framework for training LLM agents without human supervision. In this section, we first introduce agent training on synthetic tasks and the extraction of weakness signals from rollout trajectories (Section~\ref{sec:training_signal}). Then, Section~\ref{sec:signal_guided_exploration} details how these signals are used as feedback to prompt LLM-based re-exploration for new task discovery. Section~\ref{sec:task_abstraction_validation} finally describes how the discovered interactions are abstracted and validated into executable tasks and incorporated into training. The overall framework is illustrated in Fig.~\ref{fig:co_evolve}.

\subsection{Training and Signal Extraction}
\label{sec:training_signal}

\paragraph{Training on Synthetic Tasks.}
At training iteration $t$, we maintain a task set $\mathcal{D}_t$ consisting of executable synthetic tasks.
The initial task set $\mathcal{D}_0$ is obtained via unguided exploration by a large language model interacting with the environment.
As training proceeds, newly synthesized and validated tasks (described in later stages) are appended to $\mathcal{D}_t$, allowing the task distribution to evolve together with the agent.

For a task $x \in \mathcal{D}_t$, we sample a group of $K$ trajectories
$\{\tau_k\}_{k=1}^{K} \sim \pi_{\theta}(\cdot \mid x)$
and assign each trajectory a scalar reward $R(\tau_k)$.
The agent is optimized using Group Relative Policy Optimization (GRPO) \cite{guo2025deepseek} by maximizing:
\begin{equation}
\begin{split}
  \mathcal{J}(\theta)=
  & \frac{1}{\sum_{k=1}^{K} |\tau_k|} \sum_{k=1}^{K} \sum_{t=1}^{|\tau_k|} \text{CLIP}(r_{k,t}(\theta), \hat{A}_k, \epsilon) \\
  &\quad - \beta \cdot \mathbb{D}_{\text{KL}}\!\left[\pi_\theta \,\|\, \pi_{\text{ref}}\right],
\end{split}
\end{equation}
where $r_{k,t}(\theta) = \frac{\pi_{\theta}(a_t^k \mid s_t^k)}{\pi_{\theta_{\text{old}}}(a_t^k \mid s_t^k)}$ is the importance ratio, and $\text{CLIP}(r, A, \epsilon) = \min[r \cdot A, \text{clip}(r, 1-\epsilon, 1+\epsilon) \cdot A]$.
Here $\hat{A}_k$ denotes the group-relative advantage, $\pi_{\mathrm{ref}}$ is a fixed reference policy, and $\beta$ weights the KL regularization term.

\paragraph{Signal Extraction.}
Beyond policy optimization, rollout trajectories generated during training contain instances of agent underperformance. To identify such weaknesses, we analyze these trajectories and define three types of behavioral signals: forgetting signals, boundary signals, and rare signals.

\paragraph{(1) Forgetting Signals.}
Following \cite{toneva2018empirical}, we use forgetting signals to detect cases where the agent previously succeeded on a task but now fails under the current policy. Let $s_{\text{now}} \in [0,1]$ denote the task-level score of the current trajectory $\tau_{\text{now}}$, computed from the environment’s terminal reward or task-specific evaluation signal.
For each task (or task type), we maintain a sliding window of recent scores:
\[
\mathcal{H}_{\text{recent}} = \{s_{t-W+1}, \ldots, s_{t}\},
\]
where \(W\) is the window size.
A forgetting signal is triggered if
\[
\exists\, s_i \in \mathcal{H}_{\text{recent}} \text{ such that } s_i \geq 0.5 \quad \text{and} \quad s_{\text{now}} < 0.5.
\]
This condition indicates that the agent has previously succeeded on the task but now fails under the current policy. The current trajectory is marked as a forgetting signal and added to the set of signal-annotated trajectories.

\paragraph{(2) Boundary Signals.}
These signals identify tasks on which the agent exhibits high outcome variability under a fixed policy within a single training iteration. For a task \( x \in \mathcal{D}_t \), we sample a group of \( K \) trajectories  
\(\{\tau_k\}_{k=1}^{K} \sim \pi_\theta(\cdot \mid x)\), and obtain their normalized outcomes \( \tilde{R}(\tau_k) \in [0,1] \).  
A boundary signal is triggered if the sampled trajectories include both successful and failed outcomes:
\[
\exists\, \tau_i, \tau_j \text{ such that }
\tilde{R}(\tau_i) > 0.5
\quad \text{and} \quad
\tilde{R}(\tau_j) < 0.5.
\]
This condition captures tasks for which the agent’s behavior is unstable, indicating proximity to the decision boundary.  
For any task that satisfies this condition, all sampled trajectories are marked as boundary signals and added to the set of signal-annotated trajectories.

\paragraph{(3) Rare Signals.}
These are defined as action patterns that have low empirical frequency over training yet recur across multiple trajectories, indicating systematic underexploration instead of one-off stochastic events~\cite{shyalika2024comprehensive}. 
We extract an action pattern \( p \) from each trajectory and maintain its cumulative occurrence count \( c_p \).  
Let \( N \) denote the total number of observed patterns.  
When \( N \geq N_{\min} \), a rare signal is triggered if
\[
\frac{c_p}{N} < \frac{\theta}{100}
\quad \text{and} \quad
c_p > 0,
\]
where \( \theta \in (0, 100) \) is a predefined frequency threshold (e.g., \( \theta = 5 \)) that controls the rarity criterion.  All trajectories containing such patterns are marked as rare signals and added to the set of signal-annotated trajectories.
A single trajectory may trigger multiple signal types simultaneously. We evaluate forgetting, boundary, and rare signals independently and keep all activated signals because they capture complementary weaknesses.

\subsection{Signal-Guided Environment Re-exploration}
\label{sec:signal_guided_exploration}

Given the signal-annotated trajectories identified in the previous stage, we perform signal-guided environment re-exploration to collect interaction data that targets the agent’s identified weaknesses.

\paragraph{Signal-Conditioned Context Construction.}
For each signal-annotated trajectory, we provide the full interaction history to a large language model (LLM) and prompt it to reflect on the trajectory.
Each trajectory contains the task description, the agent’s executed action sequence, and the corresponding environment feedback.
Based on this information, the LLM summarizes the underlying failure cause or behavioral instability that triggered the signal and produces a structured exploration context, which characterizes where and how the agent fails or behaves unstably.
\begin{table*}[t]
    \centering
    \renewcommand{\arraystretch}{1}
    \setlength{\tabcolsep}{6.5pt}
    
    \resizebox{\linewidth}{!}{
    \begin{tabular}{lcccccc}
    \toprule
    \multirow{2}{*}{\textbf{Model}} 
    & \multicolumn{2}{c}{\textbf{AppWorld-TestN}} 
    & \multicolumn{2}{c}{\textbf{AppWorld-TestC}} 
    & \textbf{BFCL-V3} 
    & \multirow{2}{*}{\textbf{Avg.}} \\
    \cmidrule(lr){2-3}
    \cmidrule(lr){4-5}
    \cmidrule(lr){6-6}
    & \textbf{TGC} & \textbf{SGC} & \textbf{TGC} & \textbf{SGC} & \textbf{Multi-turn} & \\
    \midrule
    
    \multicolumn{7}{c}{\textit{\textbf{Closed-source LLMs}}} \\
    \midrule
    Claude-Sonnet-4.5 & 73.81 & 55.36 & 49.88 & 32.37 & 69.00 & 56.08 \\
    
    GPT-4 & 30.40 & 21.40 & 14.60 & 9.30 & 54.00 & 25.94 \\

    Gemini-2.5-Flash & 53.57 & 32.14 & 40.05 & 20.14 & 41.50 & 37.48 \\
    
    \midrule
    \multicolumn{7}{c}{\textit{\textbf{Open-source Models}}} \\
    \midrule
    DeepSeek-V3.2
& 47.02 & 30.36 & 22.78 & 9.35 & 41.50 & 30.20 \\
    
    GPT-OSS-20B
& 17.86 & 3.57 & 5.52 & 0.00 & 14.00 & 8.19 \\
    
    LLaMA-3.3-70B
& 31.54 & 8.92 & 16.31 & 4.32 & 26.00 & 17.42 \\
    
    Gemma-3-27B
& 23.81 & 7.14 & 9.83 & 0.72 & 16.50 & 11.60 \\
    
    \midrule
    \multicolumn{7}{c}{\textit{\textbf{Baseline and CoEvolve}}} \\
    \midrule
    
    Qwen2.5-7B-Instruct
& 1.19 & 0.00 & 0.72 & 0.00 & 13.50 & 3.08 \\
    
    \rowcolor[HTML]{F5F5F5}
    \textit{\textbf{-w/ CoEvolve}}
& \textbf{27.98} {\small ($\uparrow$26.79)}
& \textbf{12.50} {\small ($\uparrow$12.50)}
& \textbf{8.39}  {\small ($\uparrow$7.67)}
& \textbf{2.16}  {\small ($\uparrow$2.16)}
& \textbf{61.50} {\small ($\uparrow$48.00)}
& \textbf{22.51} {\small ($\uparrow$19.43)} \\
    
    \midrule
    Qwen3-4B-Instruct
& 16.67 & 5.36 & 7.91 & 2.16 & 26.50 & 11.72 \\
    
    \rowcolor[HTML]{F5F5F5}
    \textit{\textbf{-w/ CoEvolve}}
& \textbf{35.71} {\small ($\uparrow$19.04)}
& \textbf{14.28} {\small ($\uparrow$8.92)}
& \textbf{17.03} {\small ($\uparrow$9.12)}
& \textbf{6.47}  {\small ($\uparrow$4.31)}
& \textbf{63.00} {\small ($\uparrow$36.50)}
& \textbf{27.30} {\small ($\uparrow$15.58)} \\
    
    \midrule
    Qwen3-30B-A3B
& 31.55 & 12.50 & 19.90 & 5.76 & 43.50 & 22.64 \\
    
    \rowcolor[HTML]{F5F5F5}
    \textit{\textbf{-w/ CoEvolve}}
& \textbf{54.76} {\small ($\uparrow$23.21)}
& \textbf{33.93} {\small ($\uparrow$21.43)}
& \textbf{31.65} {\small ($\uparrow$11.75)}
& \textbf{16.55} {\small ($\uparrow$10.79)}
& \textbf{67.00} {\small ($\uparrow$23.50)}
& \textbf{40.78} {\small ($\uparrow$18.14)} \\
    
    \bottomrule
    \end{tabular}
    }
    \caption{
    Performance comparison on AppWorld (Test-Normal TGC/SGC and Test-Challenge TGC/SGC) and BFCL-V3 (Multi-turn base).
    Results are reported for closed-source LLMs, open-source models, and the backbones with and without CoEvolve.
Improvements introduced by CoEvolve are indicated by $\uparrow$.
    }
    \label{tab:main_results}
\end{table*}

\paragraph{LLM-Guided Re-exploration.}
Conditioned on the constructed context, the LLM is used to re-explore the environment to discover alternative behaviors.
For each context, exploration is conducted along two orthogonal dimensions:
(i) \emph{multi-round exploration}, where multiple independent exploration runs are initiated from the same context to encourage behavioral diversity; and
(ii) \emph{multi-step exploration}, where each exploration run proceeds for multiple interaction steps, allowing the LLM to revise its actions based on intermediate observations. During re-exploration, at each step, the LLM produces an action \(a_t\), the environment returns an observation \(o_t\), and the interaction is recorded.
As a result, the output of this stage is a collection of step-level interaction triplets \((a_t, o_t, \text{id})\), where \(\text{id}\) denotes the exploration rollout to which the step belongs. These triplets are subsequently grouped by task and serve as the input to the next stage for task abstraction and validation.

\subsection{Task Abstraction and Validation}
\label{sec:task_abstraction_validation}

Given the step-level action-observation triplets collected during the above stage, we next synthesize new executable tasks to update the task set $\mathcal{D}_t$.

\paragraph{Triplet Aggregation and Task Abstraction.}
We first group the collected interaction triplets by their associated task, where each group aggregates action-observation pairs from multiple exploration rollouts under the same task context.
These groups capture diverse behavioral evidence on how the task may be completed.
We then prompt a large language model to abstract each group into a task-level specification. Instead of copying step-level interactions, the model identifies the user intent, formulates a concise task query, and derives a plausible action sequence as a solution. This process transforms triplets into task-solution pairs.

\paragraph{Environment Validation.}
Each synthesized task-solution pair is validated through execution in the environment.
Specifically, we instantiate the environment associated with the task and provide the generated task query and action sequence to an LLM agent for execution.
If the execution successfully completes the task objective, the synthesized task is accepted.
If execution fails but the environment returns a positive reward, the task is also retained.
Tasks that fail both criteria are discarded. Validated tasks are appended to the current task set $\mathcal{D}_t$, forming the updated training distribution for the next iteration. By iteratively abstracting, validating, and incorporating new tasks, this stage allows the training data to adapt to the agent’s weaknesses, completing the co-evolution loop.

\section{Experiments}
\label{sec:experiment}

\subsection{Experimental Setup}

We evaluate our method on two widely used benchmarks: AppWorld~\cite{trivedi2024appworld} and BFCL-V3 Multi-Turn Base~\cite{patilberkeley}. 
For AppWorld, we report results on the official Test-Normal (TestN) and Test-Challenge (TestC) splits, using Task Goal Completion (TGC) and Scenario Goal Completion (SGC) to measure final task success and scenario-level execution accuracy, respectively. 
For BFCL-V3, we follow the standard Multi-Turn Base protocol and evaluate on the provided validation set, reporting multi-turn success rate.

\begin{figure*}[t]
  \includegraphics[width=\textwidth]{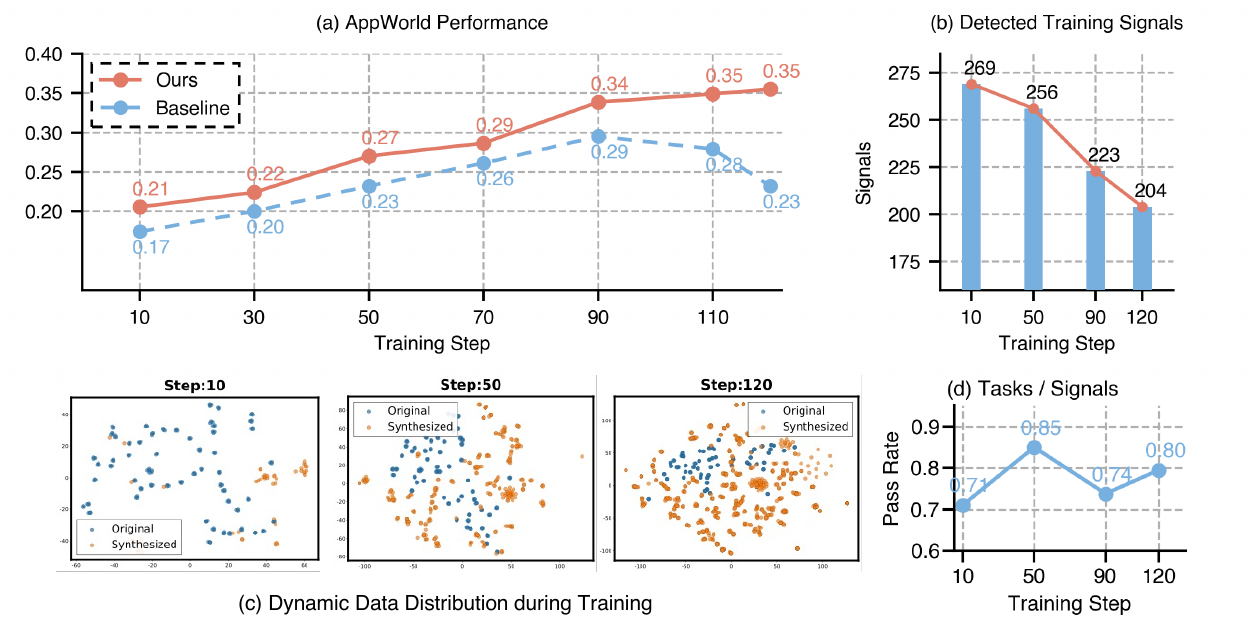}
  \caption{
Dynamics of CoEvolve during training.
(a) Performance comparison between CoEvolve and the baseline on AppWorld as training progresses.
(b) Number of detected signals across training steps.
(c) Evolution of the data distribution, showing the relationship between original and synthesized tasks at different stages.
(d) Conversion from detected weakness signals to newly generated training tasks over training.
Together, the figure shows how feedback signals guide data generation, reshape the data distribution, and support stable performance improvement.
}

  \label{fig:dynamic_train_preocess}
\end{figure*}

\begin{table*}[t]
\centering
\small
\setlength{\tabcolsep}{7pt}
\begin{tabular}{lcccccc}
\toprule
\multirow{2}{*}{\textbf{Backbone}} & \multicolumn{3}{c}{\textbf{AppWorld TestN (TGC)}} & \multicolumn{3}{c}{\textbf{BFCL-V3}} \\
\cmidrule(lr){2-4} \cmidrule(lr){5-7}
 & Zero-shot & GRPO & CoEvolve & Zero-shot & GRPO & CoEvolve \\
\midrule
Qwen2.5-7B-Instruct & 1.19 & 26.78 & \textbf{27.98} & 13.50 & 56.00 & \textbf{61.50} \\
Qwen3-4B-Instruct & 16.67 & 28.57 & \textbf{35.71} & 26.50 & 58.00 & \textbf{63.00} \\
Qwen3-30B-A3B-Instruct & 31.55 & 48.81 & \textbf{54.76} & 43.50 & 64.00 & \textbf{67.00} \\
\bottomrule
\end{tabular}
\caption{Comparison with zero-shot and GRPO on AppWorld TestN and BFCL-V3. CoEvolve is built on top of GRPO and yields complementary gains across model scales.}
\label{tab:grpo_main}
\end{table*}

\subsection{Implementation Details}
We implement all experiments with the VeRL framework \cite{wu2025hybridflow}. Specifically, Qwen2.5-7B-Instruct and Qwen3-4B-Instruct are trained on one node with 8 NVIDIA H20 GPUs,
while Qwen3-30B-A3B-Instruct is trained on 16 H20 GPUs.
We use GRPO with a constant learning rate of $\mathrm{e}{-6}$, $n{=}8$ samples per prompt, and KL coefficient $\mathrm{e}{-3}$. Rollout temperature is 0.9.

\subsection{Main Results}

Table \ref{tab:main_results} reports the main results on AppWorld and BFCL-V3, comparing closed-source LLMs, strong open-source baselines, and our backbone models trained with and without the proposed framework.

\noindent\textbf{Overall Performance.}  
CoEvolve consistently improves performance across all evaluated backbones, starting from weak instruction-following baselines. On Qwen2.5-7B, the average score increases by 19.4; Qwen3-4B improves by 15.6. These gains close the gap with much larger open-source models (e.g., DeepSeek-V3.2 at 30.20). Notably, all improvements are achieved without any human annotation or handcrafted task design, highlighting the scalability of CoEvolve as a broadly applicable training strategy rather than a model-specific trick.

\noindent\textbf{Results on AppWorld and BFCL-V3.}  
On AppWorld, CoEvolve brings +23.21 / +21.43 gains (TGC/SGC) on the challenge split and +11.75 / +10.79 on the normal split for Qwen3-30B-A3B, indicating that CoEvolve more effectively addresses failure-prone, unstable, and underexplored interaction patterns targeted by the proposed training-time feedback signals. On BFCL-V3, it improves Qwen2.5-7B-Instruct by +48.0 and Qwen3-4B-Instruct by +36.5, with smaller models benefiting more from feedback-driven training.

\noindent\textbf{Comparison with Closed-source LLMs.}  
CoEvolve enables mid-sized open models to outperform several closed-source baselines, despite lacking access to proprietary data. On BFCL-V3, Qwen3-4B with CoEvolve reaches 63.00, surpassing GPT-4 (54.00) and Gemini-2.5-Flash (41.50). These results suggest that CoEvolve improves generalization to complex interactions rather than overfitting to task environments.

\noindent\textbf{Comparison with GRPO.}  
CoEvolve extends standard GRPO by introducing feedback-guided data evolution during RL training. Table~\ref{tab:grpo_main} shows that CoEvolve consistently improves over GRPO across all three backbones,
 confirming that closed-loop data evolution provides complementary improvements on top of GRPO, rather than replacing it.

\subsection{Ablation Study}

Unless otherwise specified, all ablation experiments are conducted using the Qwen3-4B-Instruct backbone. For AppWorld, we report task-level goal completion (TGC) scores on the TestN split.

\begin{table}[t]
\centering
\small
\resizebox{\columnwidth}{!}{
\begin{tabular}{lccc}
\toprule
\textbf{Training Phase} & \textbf{AppWorld} & \textbf{BFCL} & \textbf{Avg.} \\
\midrule
Zero-shot (Qwen3-4B) & 16.67 & 26.50 & 21.59 \\
\midrule
+ Synthetic Data & 28.57 & 58.00 & 43.29 \\
+ Random Exploration & 30.36 & 60.50 & 45.43 \\
+ Feedback & \textbf{35.71} & \textbf{63.00} & \textbf{49.36} \\
\bottomrule
\end{tabular}
}
\caption{
  Ablation study of different training phases on Qwen3-4B across two benchmarks (AppWorld and BFCL). “Avg.” denotes the mean success rate across AppWorld and BFCL, showing that each phase contributes incremental gains, with the best performance achieved after incorporating feedback.
}
\label{tab:ablation_study}
\end{table}

\begin{table}[t]
\centering
\small
\resizebox{\columnwidth}{!}{%
\begin{tabular}{lccc}
\toprule
\textbf{Dimension} & \textbf{Value} & \textbf{AppWorld} & \textbf{BFCL} \\
\midrule
\multirow{3}{*}{Initial Data Size ($N$)} 
 & 50  & {26.79} & {53.00} \\
 & 100 & {35.12}    & \textbf{63.00}    \\
 & 200 & \textbf{38.10} & {60.00}    \\
\midrule
\multirow{3}{*}{Gen. Frequency ($F$)} 
 & 5  & \textbf{35.71} & {58.00} \\
 & 10 & {35.12}    & \textbf{63.00} \\
 & 20 & {33.93} & {57.50} \\
\bottomrule
\end{tabular}
}
\caption{Hyperparameter sensitivity analysis for Qwen3-4B on AppWorld and BFCL benchmarks. We investigate the impact of initial synthetic data size ($N$), and generation frequency ($F$).}
\label{tab:hyperparameter_analysis}
\end{table}

\begin{figure}[t]
  \includegraphics[width=\columnwidth]{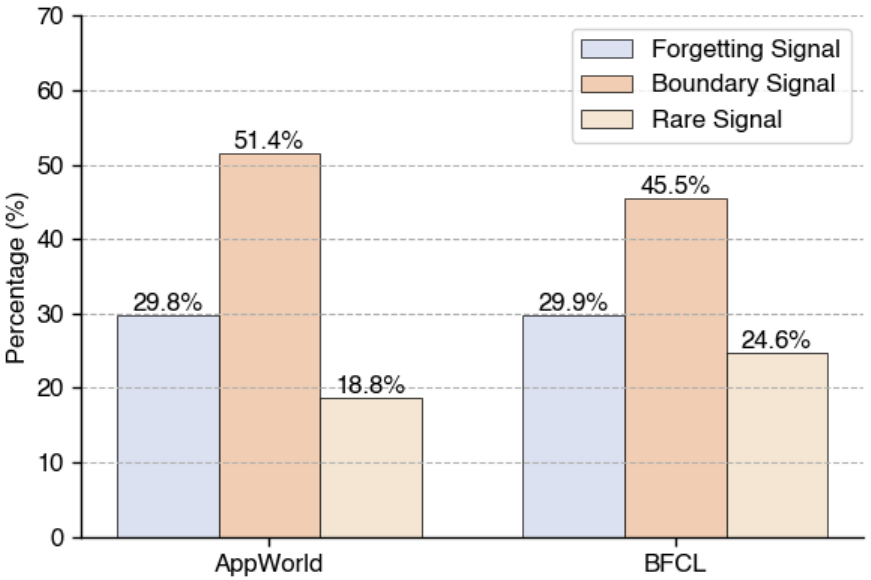}
  \caption{
Distribution of extracted signals on AppWorld and BFCL. Boundary signals dominate (51.4\% on AppWorld and 45.5\% on BFCL), followed by forgetting and rare signals.
}

  \label{fig:signal_distribution}
\end{figure}

\noindent\textbf{Dynamics of Agent--Data CoEvolve.}  
Fig.~\ref{fig:dynamic_train_preocess} shows how agent performance, detected signals, and synthesized data evolve throughout training. Performance improves steadily (0.21 $\rightarrow$ 0.35), while the baseline rises initially before falling (0.17 $\rightarrow$ 0.29 $\rightarrow$ 0.23), indicating more stable optimization under closed-loop training. Generated tasks expand into previously underrepresented regions (Fig.~\ref{fig:dynamic_train_preocess}(c)), showing that synthesis produces diverse, non-redundant data. The number of detected signals drops from 269 to 204 (Fig.~\ref{fig:dynamic_train_preocess}(b)), suggesting progressive resolution of failure-prone cases. The pass rate of signal-driven tasks improves from 0.71 to 0.85 and stabilizes at 0.80 (Fig.~\ref{fig:dynamic_train_preocess}(d)), confirming the effectiveness of feedback-guided generation. Overall, these trends support CoEvolve’s core design: using feedback signals to adaptively reshape data distribution and target evolving model weaknesses.

\noindent\textbf{Effect of Closed-loop CoEvolve.}  
Table~\ref{tab:ablation_study} shows an ablation study on Qwen3-4B, isolating the impact of each training stage. Starting from a zero-shot baseline (21.59), static synthetic data already provides a strong boost (43.29), confirming the value of offline task construction. Adding random exploration brings further gains (45.43), indicating that online trajectory generation can help. However, the most significant improvement comes from incorporating feedback signals, which raises the average score to 49.36. Compared with random exploration, feedback-guided generation yields consistent gains on AppWorld (30.36 $\rightarrow$ 35.71) and BFCL-V3 (60.50 $\rightarrow$ 63.00), underscoring the importance of using model feedback to shape the evolving training set.

\noindent\textbf{Hyperparameter Sensitivity.}  
Table~\ref{tab:hyperparameter_analysis} shows that both initialization size ($N$) and generation frequency ($F$) affect performance. $N=100$ gives the best BFCL score (63.00), while $N=200$ improves AppWorld (38.10) but slightly degrades BFCL. For $F$, $F=5$ achieves the highest AppWorld score (35.71), while $F=10$ yields the best BFCL score (63.00). Overly sparse updates ($F=20$) degrade both. These results suggest that moderate initialization and sufficiently frequent updates are important for feedback-driven training.

\noindent\textbf{Ablation and Distribution of Feedback Signals.}  
To better understand the role of individual feedback signals, we analyze both their distribution and their impact on performance. As shown in Figure~\ref{fig:signal_distribution}, boundary signals account for the largest proportion across both AppWorld (51.4\%) and BFCL (45.5\%), followed by forgetting and rare signals. This suggests that agents frequently struggle at decision boundaries and with previously learned cases, justifying their use as guidance for task synthesis. Table~\ref{tab:feedback_ablation} further shows that removing any single signal leads to performance degradation, confirming their complementary value. In particular, forgetting signals contribute the most, with a drop of nearly 4 points (49.36 $\rightarrow$ 45.18), reflecting their utility in correcting regressions during training. Boundary and rare signals also provide meaningful gains (1.6$\sim$1.9), indicating their importance in exposing edge cases and long-tail scenarios. Together, these results validate that CoEvolve benefits from a diverse signal set rather than a single heuristic.

\begin{table}[t]
\centering
\small
\resizebox{\columnwidth}{!}{%
\begin{tabular}{lccc}
\toprule
\textbf{Training Configuration} & \textbf{AppWorld} & \textbf{BFCL} & \textbf{Avg.} \\
\midrule
CoEvolve & \textbf{35.71} & \textbf{63.00} & \textbf{49.36} \\
\midrule
\quad w/o Forgotten Signals & 30.36  & 60.00 & 45.18 \\
\quad w/o Rare Signals & 33.92 & 60.50 & 47.21 \\
\quad w/o Boundary Signals & 33.33 & 61.00 & 47.17
 \\
\bottomrule
\end{tabular}
}

\caption{Ablation study on feedback signals for Qwen3-4B. Each row represents the performance after removing a specific type of feedback-driven sample.}
\label{tab:feedback_ablation}
\end{table}

\begin{table}[t]
\centering
\small
\resizebox{\columnwidth}{!}{%
\begin{tabular}{lcc}
\toprule
\textbf{Training Domain} & \textbf{AppWorld} & \textbf{BFCL} \\
\midrule
Zero-shot Baseline (Qwen3-4B) & 16.67 & 26.50 \\
\midrule
AppWorld (Ours) & \textbf{35.71} & 45.00 \\
BFCL (Ours) & 19.04 & \textbf{63.00} \\
\bottomrule
\end{tabular}
}
\caption{Cross-domain transferability analysis of Qwen3-4B. The diagonal entries represent intra-domain performance, while off-diagonal entries indicate zero-shot generalization to unseen tool-use environments.}
\label{tab:cross_domain}
\end{table}

\begin{table*}[t]
\centering
\small
\setlength{\tabcolsep}{7pt}
\renewcommand{\arraystretch}{1.15}
\begin{tabular}{lcccccc}
\toprule
\textbf{Benchmark} 
& \textbf{Feedback Time Ratio} 
& \textbf{Single Feedback Cost} 
& \textbf{Total Time} 
& \textbf{Performance Gain} 
& \textbf{Relative Gain} \\
\midrule
AppWorld
& 9.67\% 
& 811s 
& 100{,}596s 
& 28.57 $\rightarrow$ 35.12
& +22.92\% \\
\midrule
BFCL 
& 12.76\% 
& 480s 
& 45{,}144s 
& 58.00 $\rightarrow$ 63.00 
& +8.62\% \\
\bottomrule
\end{tabular}
\caption{Cost and Efficiency of CoEvolve.
CoEvolve introduces minimal computational overhead, yet yields substantial performance gains across benchmarks, showing its effectiveness as an efficient training strategy.}
\label{tab:feedback_cost_efficiency}
\end{table*}

\begin{table}[t]
    \renewcommand{\arraystretch}{1.2}

    \footnotesize
    \centering
    \setlength{\tabcolsep}{2.2mm}{
        \begin{tabular}{c|c|c c}
            \toprule
            Benchmark & \diagbox{Baseline}{Ours} & Correct & Wrong \\
            \midrule
            \multirow{2}{*}{BFCL}
            & Correct & 53.00\% &  5.00\%\\
            & Wrong   &  10.00\% & 32.00\% \\
            \midrule
            \multirow{2}{*}{AppWorld}
            & Correct & 19.04\% &  9.53\%\\
            & Wrong   & 16.67\%  & 54.76\% \\
            \bottomrule
        \end{tabular}
        \caption{Cross comparison between CoEvolve (Ours) and the baseline on BFCL and AppWorld.}
        \label{tab:cross_comparison}
    }

\end{table}

\noindent\textbf{Cross-domain Generalization.}  
Table~\ref{tab:cross_domain} evaluates whether CoEvolve-trained agents generalize across domains. Training on AppWorld improves zero-shot performance on BFCL from 26.50 to 45.00, and vice versa from 16.67 to 19.04. While in-domain performance remains highest (35.71, 63.00), the off-domain gains show that CoEvolve learns transferable strategies beyond environment-specific behaviors.

\begin{figure}[t]
  \includegraphics[width=\columnwidth]{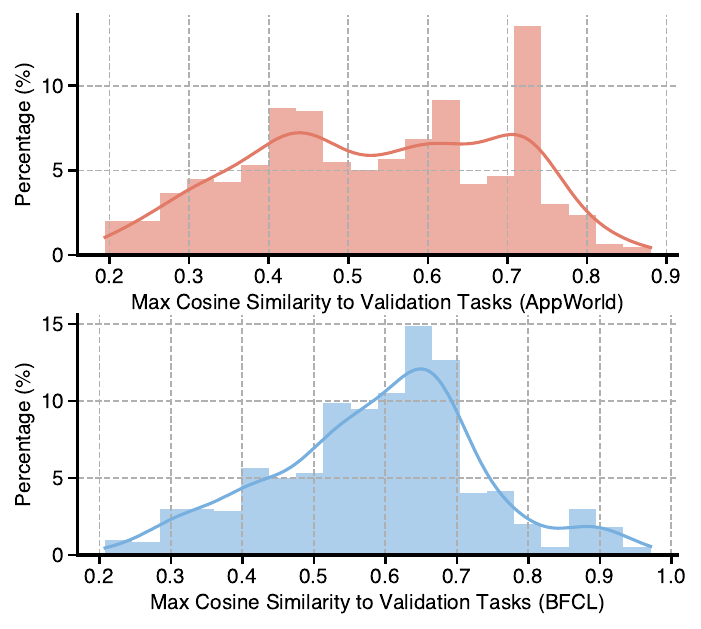}
  \caption{
Distribution of maximum cosine similarity between synthesized tasks and their validation tasks.
}

  \label{fig:similarity_distribution}
\end{figure}

\noindent\textbf{Analysis of Data Diversity.}  
Fig.~\ref{fig:similarity_distribution} analyzes the similarity between synthesized tasks and validation examples. Across both AppWorld and BFCL, most samples fall within a moderate similarity range (e.g., 0.4$\sim$0.7), with only a small fraction approaching 1.0. This indicates that the synthesis process produces novel tasks rather than near-duplicates. The consistent patterns observed across domains further suggest that the feedback-driven exploration effectively guides task discovery, maintaining meaningful data diversity without collapsing into repetitive samples.

\noindent\textbf{Behavioral Comparison with GRPO Baseline.}
Table~\ref{tab:cross_comparison} compares CoEvolve against a GRPO-trained baseline without closed-loop evolution. On BFCL, CoEvolve preserves 53.00\% of correct predictions and recovers 10.00\% of previously failed cases. On AppWorld, the corresponding numbers are 19.04\% and 16.67\%. These results indicate that feedback-driven training not only retains prior strengths but also effectively corrects failure cases, yielding more reliable and adaptive agent behavior.

\noindent\textbf{Cost-Efficiency of CoEvolve.}  
Table~\ref{tab:feedback_cost_efficiency} evaluates the additional training cost and corresponding performance improvement brought by CoEvolve, compared to a baseline that performs GRPO training without closed-loop task generation. Across both benchmarks, the CoEvolve framework introduces only $\sim$10\% additional computation, yet leads to clear absolute gains (+6.53 on AppWorld, +5.00 on BFCL) and substantial relative improvements (+22.92\%, +8.62\%). Each feedback iteration incurs minimal cost, but collectively reshapes the training distribution to better address model weaknesses. These results confirm that CoEvolve is not only effective but also efficient, offering a favorable trade-off between cost and performance.

\section{Conclusion}

We introduce CoEvolve, a reinforcement learning framework that enables mutual evolution between the agent and its data distribution. By extracting feedback signals (e.g., forgetting signal) during policy optimization and using them to guide task synthesis, our method progressively adapts both the agent's capabilities and the data it learns from. Extensive experiments on AppWorld and BFCL validate its effectiveness and efficiency. We hope this work inspires future research on agent evolution toward agents that can autonomously improve via interaction-driven feedback.
\section*{Limitations}

This work presents an exploration of feedback-driven agent-data co-evolution using a limited set of feedback signals, including forgetting signals, boundary signals, and rare signals. While effective, these signals cover only a subset of potentially informative feedback and may be further enriched in future work. In addition, the extracted signals are derived from the agent’s own interaction trajectories and therefore depend on the current policy. At early stages of training, when the agent’s behavior is still immature, the resulting signals may be noisy or incomplete, highlighting the need for more robust feedback extraction under low-competence regimes.
Because CoEvolve autonomously reshapes its training distribution, adversarial or safety-critical settings may require human oversight, policy constraints, and continuous auditing before synthesized tasks are admitted into training. Future work should incorporate explicit safety filters and risk-triggered review so that feedback-driven adaptation remains controllable.

\bibliography{custom}

@article{jin2024llms,
  title={From llms to llm-based agents for software engineering: A survey of current, challenges and future},
  author={Jin, Haolin and Huang, Linghan and Cai, Haipeng and Yan, Jun and Li, Bo and Chen, Huaming},
  journal={arXiv preprint arXiv:2408.02479},
  year={2024}
}

@inproceedings{ding2025toolcoder,
  title={Toolcoder: A systematic code-empowered tool learning framework for large language models},
  author={Ding, Hanxing and Tao, Shuchang and Pang, Liang and Wei, Zihao and Gao, Jinyang and Ding, Bolin and Shen, Huawei and Cheng, Xueqi},
  booktitle={Proceedings of the 63rd Annual Meeting of the Association for Computational Linguistics (Volume 1: Long Papers)},
  pages={17876--17891},
  year={2025}
}

@article{wang2025co,
  title={Co-evolving llm coder and unit tester via reinforcement learning},
  author={Wang, Yinjie and Yang, Ling and Tian, Ye and Shen, Ke and Wang, Mengdi},
  journal={arXiv preprint arXiv:2506.03136},
  year={2025}
}

@article{gou2025empirical,
  title={An empirical study on how video-LLMs answer video questions},
  author={Gou, Chenhui and Ma, Ziyu and Duan, Zicheng and He, Haoyu and Chen, Feng and Liu, Akide and Zhuang, Bohan and Cai, Jianfei and Rezatofighi, Hamid},
  journal={arXiv preprint arXiv:2508.15360},
  year={2025}
}

@article{ma2024drvideo,
  title={Drvideo: Document retrieval based long video understanding},
  author={Ma, Ziyu and Gou, Chenhui and Shi, Hengcan and Sun, Bin and Li, Shutao and Rezatofighi, Hamid and Cai, Jianfei},
  journal={arXiv preprint arXiv:2406.12846},
  year={2024}
}

@article{Ma_Gou_Hu_Wang_Zhuang_Cai_2026, title={Where and What Matters: Sensitivity-Aware Task Vectors for Many-Shot Multimodal In-Context Learning}, volume={40}, url={https://ojs.aaai.org/index.php/AAAI/article/view/37733}, DOI={10.1609/aaai.v40i10.37733}, number={10}, journal={Proceedings of the AAAI Conference on Artificial Intelligence}, author={Ma, Ziyu and Gou, Chenhui and Hu, Yiming and Wang, Yong and Zhuang, Bohan and Cai, Jianfei}, year={2026}, month={Mar.}, pages={7892-7900} }

@article{ji2025tree,
  title={Tree search for llm agent reinforcement learning},
  author={Ji, Yuxiang and Ma, Ziyu and Wang, Yong and Chen, Guanhua and Chu, Xiangxiang and Wu, Liaoni},
  journal={arXiv preprint arXiv:2509.21240},
  year={2025}
}

@article{sun2024llm,
  title={Llm-based multi-agent reinforcement learning: Current and future directions},
  author={Sun, Chuanneng and Huang, Songjun and Pompili, Dario},
  journal={arXiv preprint arXiv:2405.11106},
  year={2024}
}

@article{guo2025deepseek,
  title={DeepSeek-R1 incentivizes reasoning in LLMs through reinforcement learning},
  author={Guo, Daya and Yang, Dejian and Zhang, Haowei and Song, Junxiao and Wang, Peiyi and Zhu, Qihao and Xu, Runxin and Zhang, Ruoyu and Ma, Shirong and Bi, Xiao and others},
  journal={Nature},
  volume={645},
  number={8081},
  pages={633--638},
  year={2025},
  publisher={Nature Publishing Group UK London}
}

@article{zhai2025agentevolver,
  title={AgentEvolver: Towards Efficient Self-Evolving Agent System},
  author={Zhai, Yunpeng and Tao, Shuchang and Chen, Cheng and Zou, Anni and Chen, Ziqian and Fu, Qingxu and Mai, Shinji and Yu, Li and Deng, Jiaji and Cao, Zouying and others},
  journal={arXiv preprint arXiv:2511.10395},
  year={2025}
}

@inproceedings{gur2023understanding,
  title={Understanding html with large language models},
  author={G{\"u}r, Izzeddin and Nachum, Ofir and Miao, Yingjie and Safdari, Mustafa and Huang, Austin and Chowdhery, Aakanksha and Narang, Sharan and Fiedel, Noah and Faust, Aleksandra},
  booktitle={Findings of the Association for Computational Linguistics: EMNLP 2023},
  pages={2803--2821},
  year={2023}
}

@article{li2025deepagent,
  title={Deepagent: A general reasoning agent with scalable toolsets},
  author={Li, Xiaoxi and Jiao, Wenxiang and Jin, Jiarui and Dong, Guanting and Jin, Jiajie and Wang, Yinuo and Wang, Hao and Zhu, Yutao and Wen, Ji-Rong and Lu, Yuan and others},
  journal={arXiv preprint arXiv:2510.21618},
  year={2025}
}

@article{mai2025cues,
  title={CuES: A Curiosity-driven and Environment-grounded Synthesis Framework for Agentic RL},
  author={Mai, Shinji and Zhai, Yunpeng and Chen, Ziqian and Chen, Cheng and Zou, Anni and Tao, Shuchang and Liu, Zhaoyang and Ding, Bolin},
  journal={arXiv preprint arXiv:2512.01311},
  year={2025}
}

@article{achiam2023gpt,
  title   = {GPT-4 Technical Report},
  author  = {Achiam, Josh and Adler, Steven and Agarwal, Sandhini and Ahmad, Lama and Akkaya, Ilge and Aleman, Florencia Leoni and Almeida, Diogo and Altenschmidt, Janko and Altman, Sam and Anadkat, Shyamal and others},
  journal = {arXiv preprint arXiv:2303.08774},
  year    = {2023}
}

@article{comanici2025gemini,
  title={Gemini 2.5: Pushing the frontier with advanced reasoning, multimodality, long context, and next generation agentic capabilities},
  author={Comanici, Gheorghe and Bieber, Eric and Schaekermann, Mike and Pasupat, Ice and Sachdeva, Noveen and Dhillon, Inderjit and Blistein, Marcel and Ram, Ori and Zhang, Dan and Rosen, Evan and others},
  journal={arXiv preprint arXiv:2507.06261},
  year={2025}
}

@article{grattafiori2024llama,
  title   = {The LLaMA 3 Herd of Models},
  author={Grattafiori, Aaron and Dubey, Abhimanyu and Jauhri, Abhinav and Pandey, Abhinav and Kadian, Abhishek and Al-Dahle, Ahmad and Letman, Aiesha and Mathur, Akhil and Schelten, Alan and Vaughan, Alex and others},
  journal={arXiv preprint arXiv:2407.21783},
  year={2024}
}

@article{lin2025comprehensive,
  title={A comprehensive survey on reinforcement learning-based agentic search: Foundations, roles, optimizations, evaluations, and applications},
  author={Lin, Minhua and Wu, Zongyu and Xu, Zhichao and Liu, Hui and Tang, Xianfeng and He, Qi and Aggarwal, Charu and Zhang, Xiang and Wang, Suhang},
  journal={arXiv preprint arXiv:2510.16724},
  year={2025}
}

@article{liu2024deepseek,
  title   = {DeepSeek-V3 Technical Report},
  author  = {Liu, Aixin and Feng, Bei and Xue, Bing and Wang, Bingxuan and Wu, Bochao and Lu, Chengda and Zhao, Chenggang and Deng, Chengqi and Zhang, Chenyu and Ruan, Chong and others},
  journal = {arXiv preprint arXiv:2412.19437},
  year    = {2024}
}

@article{mai2025agent,
  title={Agent rl scaling law: Agent rl with spontaneous code execution for mathematical problem solving},
  author={Mai, Xinji and Xu, Haotian and Li, Zhong-Zhi and W, Xing and Wang, Weinong and Hu, Jian and Zhang, Yingying and Zhang, Wenqiang},
  journal={arXiv preprint arXiv:2505.07773},
  year={2025}
}

@article{yu2025demystifying,
  title   = {Demystifying Reinforcement Learning in Agentic Reasoning},
  author  = {Yu, Zhaochen and Yang, Ling and Zou, Jiaru and Yan, Shuicheng and Wang, Mengdi},
  journal = {arXiv preprint arXiv:2510.11701},
  year    = {2025}
}

@inproceedings{trivedi2024appworld,
  title={Appworld: A controllable world of apps and people for benchmarking interactive coding agents},
  author={Trivedi, Harsh and Khot, Tushar and Hartmann, Mareike and Manku, Ruskin and Dong, Vinty and Li, Edward and Gupta, Shashank and Sabharwal, Ashish and Balasubramanian, Niranjan},
  booktitle={Proceedings of the 62nd Annual Meeting of the Association for Computational Linguistics (Volume 1: Long Papers)},
  pages={16022--16076},
  year={2024}
}

@inproceedings{patilberkeley,
  title={The Berkeley Function Calling Leaderboard (BFCL): From Tool Use to Agentic Evaluation of Large Language Models},
  author={Patil, Shishir G and Mao, Huanzhi and Yan, Fanjia and Ji, Charlie Cheng-Jie and Suresh, Vishnu and Stoica, Ion and Gonzalez, Joseph E},
  booktitle={Forty-second International Conference on Machine Learning},
  year={2025}
}

@misc{nakano2021,
  title         = {WebGPT: Browser-assisted Question-Answering with Human Feedback},
  author        = {Reiichiro Nakano and Jacob Hilton and Suchir Balaji and Jeff Wu and Long Ouyang and Christina Kim and Christopher Hesse and Shantanu Jain and Vineet Kosaraju and William Saunders and Xu Jiang and Karl Cobbe and Tyna Eloundou and Gretchen Krueger and Kevin Button and Matthew Knight and Benjamin Chess and John Schulman},
  year          = {2021},
  eprint        = {2112.09332},
  archivePrefix = {arXiv},
  primaryClass  = {cs.CL}
}

@inproceedings{zhang2024knowagent,
  title={Knowagent: Knowledge-augmented planning for llm-based agents},
  author={Zhu, Yuqi and Qiao, Shuofei and Ou, Yixin and Deng, Shumin and Lyu, Shiwei and Shen, Yue and Liang, Lei and Gu, Jinjie and Chen, Huajun and Zhang, Ningyu},
  booktitle={Findings of the Association for Computational Linguistics: NAACL 2025},
  pages={3709--3732},
  year={2025}
}

@article{ye2024,
  title   = {{LLM-DA}: Data Augmentation via Large Language Models for Few-Shot Named Entity Recognition},
  author  = {Ye, Junjie and Gao, Xuanteng and Zhang, Keqing and Gao, Guodian and Li, Yutao and Liu, Xiaozhi},
  journal = {arXiv preprint arXiv:2402.14568},
  year    = {2024}
}

@inproceedings{ding2024,
  title={Data augmentation using llms: Data perspectives, learning paradigms and challenges},
  author={Ding, Bosheng and Qin, Chengwei and Zhao, Ruochen and Luo, Tianze and Li, Xinze and Chen, Guizhen and Xia, Wenhan and Hu, Junjie and Tuan, Luu Anh and Joty, Shafiq},
  booktitle={Findings of the Association for Computational Linguistics: ACL 2024},
  pages={1679--1705},
  year={2024}
}

@article{wang2023,
  title={Voyager: An open-ended embodied agent with large language models},
  author={Wang, Guanzhi and Xie, Yuqi and Jiang, Yunfan and Mandlekar, Ajay and Xiao, Chaowei and Zhu, Yuke and Fan, Linxi and Anandkumar, Anima},
  journal={arXiv preprint arXiv:2305.16291},
  year={2023}
}

@article{chen2025,
  title={Training LLM-Based Agents with Synthetic Self-Reflected Trajectories and Partial Masking},
  author={Chen, Yihan and Xu, Benfeng and Wang, Xiaorui and Zhang, Yongdong and Mao, Zhendong},
  journal={arXiv preprint arXiv:2505.20023},
  year={2025}
}

@inproceedings{pahuja2025,
  title={Explorer: Scaling exploration-driven web trajectory synthesis for multimodal web agents},
  author={Pahuja, Vardaan and Lu, Yadong and Rosset, Corby and Gou, Boyu and Mitra, Arindam and Whitehead, Spencer and Su, Yu and Hassan, Ahmed},
  booktitle={Findings of the Association for Computational Linguistics: ACL 2025},
  pages={6300--6323},
  year={2025}
}

@article{yuan2024,
  title={AgentTrek: Agent Trajectory Synthesis via Guiding Replay with Web Tutorials},
  author={Xu, Yiheng and Lu, Dunjie and Shen, Zhennan and Wang, Junli and Wang, Zekun and Mao, Yuchen and Xiong, Caiming and Yu, Tao},
  journal={arXiv preprint arXiv:2412.09605},
  year={2024}
}

@article{yuan2025,
  title   = {Agent-{R}: Training Language Model Agents to Reflect via Iterative Self-Training},
  author  = {Yuan, Siyu and Chen, Zehui and Xi, Zhiheng and Ye, Junjie and Du, Zhengyin and Chen, Jiecao},
  journal = {arXiv preprint arXiv:2501.11425},
  year    = {2025}
}

@inproceedings{zhang2025,
  title={Agentbank: Towards generalized llm agents via fine-tuning on 50000+ interaction trajectories},
  author={Song, Yifan and Xiong, Weimin and Zhao, Xiutian and Zhu, Dawei and Wu, Wenhao and Wang, Ke and Li, Cheng and Peng, Wei and Li, Sujian},
  booktitle={Findings of the Association for Computational Linguistics: EMNLP 2024},
  pages={2124--2141},
  year={2024}
}

@inproceedings{luo2025,
    title = "{ST}e{C}a: Step-level Trajectory Calibration for {LLM} Agent Learning",
    author = "Wang, Hanlin  and
      Wang, Jian  and
      Leong, Chak Tou  and
      Li, Wenjie",
    editor = "Che, Wanxiang  and
      Nabende, Joyce  and
      Shutova, Ekaterina  and
      Pilehvar, Mohammad Taher",
    booktitle = "Findings of the Association for Computational Linguistics: ACL 2025",
    month = jul,
    year = "2025",
    address = "Vienna, Austria",
    publisher = "Association for Computational Linguistics",
    url = "https://aclanthology.org/2025.findings-acl.604/",
    doi = "10.18653/v1/2025.findings-acl.604",
    pages = "11597--11614",
    ISBN = "979-8-89176-256-5"
}

@article{hoang2025,
  title   = {{LAM SIMULATOR}: Advancing Data Generation for Large Action Model Training via Online Exploration and Trajectory Feedback},
  author  = {Hoang, Thai Quoc and Huang, Kung-Hsiang and Kokane, Shirley and Zhang, Jianguo and Liu, Zuxin and Zhu, Ming and Grigsby, Jake and Lan, Tian and Ryoo, Michael S and Wu, Chien-Sheng and Heinecke, Shelby and Wang, Huan and Savarese, Silvio and Xiong, Caiming and Niebles, Juan Carlos},
  journal = {Findings of the Association for Computational Linguistics: ACL 2025},
  year    = {2025}
}

@article{wang2025uisimulator,
  title={Llms as scalable, general-purpose simulators for evolving digital agent training},
  author={Wang, Yiming and Yin, Da and Cui, Yuedong and Zheng, Ruichen and Li, Zhiqian and Lin, Zongyu and Wu, Di and Wu, Xueqing and Ye, Chenchen and Zhou, Yu and others},
  journal={arXiv preprint arXiv:2510.14969},
  year={2025}
}

@article{zhao2025autoplay,
  title={Scaling synthetic task generation for agents via exploration},
  author={Ramrakhya, Ram and Szot, Andrew and Attia, Omar and Yang, Yuhao and Nguyen, Anh and Mazoure, Bogdan and Gan, Zhe and Agrawal, Harsh and Toshev, Alexander},
  journal={arXiv preprint arXiv:2509.25047},
  year={2025}
}

@article{zhang2025deepanalyze,
  title={Deepanalyze: Agentic Large Language Models for Autonomous Data Science},
  author={Zhang, Shaolei and Fan, Ju and Fan, Meihao and Li, Guoliang and Du, Xiaoyong},
  journal={arXiv preprint arXiv:2510.16872},
  year={2025}
}

@article{liu2025limi,
  title={Limi: Less is more for agency},
  author={Xiao, Yang and Jiang, Mohan and Sun, Jie and Li, Keyu and Lin, Jifan and Zhuang, Yumin and Zeng, Ji and Xia, Shijie and Hua, Qishuo and Li, Xuefeng and others},
  journal={arXiv preprint arXiv:2509.17567},
  year={2025}
}

@article{sun2025earlyexperience,
  title={Agent learning via early experience},
  author={Zhang, Kai and Chen, Xiangchao and Liu, Bo and Xue, Tianci and Liao, Zeyi and Liu, Zhihan and Wang, Xiyao and Ning, Yuting and Chen, Zhaorun and Fu, Xiaohan and others},
  journal={arXiv preprint arXiv:2510.08558},
  year={2025}
}

@inproceedings{shridhar2020alfworld,
  title     = {{ALFWorld}: Aligning Text and Embodied Environments for Interactive Learning},
  author    = {Shridhar, Mohit and Yuan, Xingdi and C{\^o}t{\'e}, Marc-Alexandre and Bisk, Yonatan and Trischler, Adam and Hausknecht, Matthew},
  booktitle = {International Conference on Learning Representations (ICLR)},
  year      = {2021},
  url       = {https://arxiv.org/abs/2010.03768}
}

@article{yang2025qwen3,
  title   = {{Qwen3} Technical Report},
  author  = {Qwen},
  journal = {arXiv preprint arXiv:2505.09388},
  year    = {2025},
  url     = {https://arxiv.org/abs/2505.09388}
}

@article{qwen25,
  author       = {Qwen},
  title        = {{Qwen2.5} Technical Report},
  journal      = {arXiv preprint arXiv:2412.15115},
  url = {https://arxiv.org/abs/2412.15115},
  year         = {2024},
}

@article{chen2025stepwise,
  title={Stepwise guided policy optimization: Coloring your incorrect reasoning in grpo},
  author={Chen, Peter and Li, Xiaopeng and Li, Ziniu and Chen, Xi and Lin, Tianyi},
  journal={arXiv preprint arXiv:2505.11595},
  year={2025}
}

@article{chen2025compo,
  title={Compo: Preference alignment via comparison oracles},
  author={Chen, Peter and Chen, Xi and Yin, Wotao and Lin, Tianyi},
  journal={arXiv preprint arXiv:2505.05465},
  year={2025}
}

@misc{anthropic2025sonnet45,
  title        = {Claude Sonnet 4.5},
  author       = {{Anthropic}},
  year         = {2025},
  howpublished = {\url{https://docs.anthropic.com/claude/docs/models-overview}},
  note         = {Accessed: 2025-01}
}

@article{liu2025deepseek,
  title={Deepseek-v3. 2: Pushing the frontier of open large language models},
  author={Liu, Aixin and Mei, Aoxue and Lin, Bangcai and Xue, Bing and Wang, Bingxuan and Xu, Bingzheng and Wu, Bochao and Zhang, Bowei and Lin, Chaofan and Dong, Chen and others},
  journal={arXiv preprint arXiv:2512.02556},
  year={2025}
}

@article{agarwal2025gpt,
  title={gpt-oss-120b \& gpt-oss-20b model card},
  author={Agarwal, Sandhini and Ahmad, Lama and Ai, Jason and Altman, Sam and Applebaum, Andy and Arbus, Edwin and Arora, Rahul K and Bai, Yu and Baker, Bowen and Bao, Haiming and others},
  journal={arXiv preprint arXiv:2508.10925},
  year={2025}
}

@article{team2025gemma,
  title={Gemma 3 technical report},
  author={Kamath, Aishwarya and Ferret, Johan and Pathak, Shreya and Vieillard, Nino and Merhej, Ramona and Perrin, Sarah and Matejovicova, Tatiana and Ram{\'e}, Alexandre and Rivi{\`e}re, Morgane and others},
  journal={arXiv preprint arXiv:2503.19786},
  year={2025}
}

@article{team2024qwen2,
  title={Qwen2 technical report},
  author={Qwen},
  journal={arXiv preprint arXiv:2407.10671},
  volume={2},
  number={3},
  year={2024}
}

@article{shinn2023reflexion,
  title={Reflexion: Language agents with verbal reinforcement learning},
  author={Shinn, Noah and Cassano, Federico and Gopinath, Ashwin and Narasimhan, Karthik and Yao, Shunyu},
  journal={Advances in Neural Information Processing Systems},
  volume={36},
  pages={8634--8652},
  year={2023}
}

@article{gulcehre2023reinforced,
  title={Reinforced self-training (rest) for language modeling},
  author={Gulcehre, Caglar and Paine, Tom Le and Srinivasan, Srivatsan and Konyushkova, Ksenia and Weerts, Lotte and Sharma, Abhishek and Siddhant, Aditya and Ahern, Alex and Wang, Miaosen and Gu, Chenjie and others},
  journal={arXiv preprint arXiv:2308.08998},
  year={2023}
}

@inproceedings{yao2022react,
  title={React: Synergizing reasoning and acting in language models},
  author={Yao, Shunyu and Zhao, Jeffrey and Yu, Dian and Du, Nan and Shafran, Izhak and Narasimhan, Karthik R and Cao, Yuan},
  booktitle={The Eleventh International Conference on Learning Representations},
  year={2023}
}

@article{chu2025gpg,
  title={Gpg: A simple and strong reinforcement learning baseline for model reasoning},
  author={Chu, Xiangxiang and Huang, Hailang and Zhang, Xiao and Wei, Fei and Wang, Yong},
  journal={arXiv preprint arXiv:2504.02546},
  year={2025}
}

@article{toneva2018empirical,
  title={An empirical study of example forgetting during deep neural network learning},
  author={Toneva, Mariya and Sordoni, Alessandro and Combes, Remi Tachet des and Trischler, Adam and Bengio, Yoshua and Gordon, Geoffrey J},
  journal={arXiv preprint arXiv:1812.05159},
  year={2018}
}

@article{shyalika2024comprehensive,
  title={A comprehensive survey on rare event prediction},
  author={Shyalika, Chathurangi and Wickramarachchi, Ruwan and Sheth, Amit P},
  journal={ACM Computing Surveys},
  volume={57},
  number={3},
  pages={1--39},
  year={2024},
  publisher={ACM New York, NY}
}

@misc{qwen3max,
    title = {Qwen3-Max: Just Scale it},
    author = {Qwen Team},
    month = {September},
    year = {2025}
}

@article{yao2022webshop,
  title={Webshop: Towards scalable real-world web interaction with grounded language agents},
  author={Yao, Shunyu and Chen, Howard and Yang, John and Narasimhan, Karthik},
  journal={Advances in Neural Information Processing Systems},
  volume={35},
  pages={20744--20757},
  year={2022}
}

@inproceedings{wu2025hybridflow,
  title={Hybridflow: A flexible and efficient rlhf framework},
  author={Sheng, Guangming and Zhang, Chi and Ye, Zilingfeng and Wu, Xibin and Zhang, Wang and Zhang, Ru and Peng, Yanghua and Lin, Haibin and Wu, Chuan},
  booktitle={Proceedings of the Twentieth European Conference on Computer Systems},
  pages={1279--1297},
  year={2025}
}

\appendix

\section{Appendix}
\label{sec:appendix}

\subsection{Dataset}
\label{sec:appendix:datasets}

For completeness, we summarize the datasets used in both the main paper and the appendix-only transfer experiments. Besides AppWorld and BFCL, the appendix additionally reports results on ALFWorld and WebShop to assess transfer beyond API-centric settings.

\paragraph{AppWorld.}
AppWorld is a simulated environment for real-world digital service interactions, covering applications such as calendar, email, music, and social platforms.
Agents solve tasks via Python API calls (e.g., ``find the most-liked song in my Spotify playlists''), typically requiring multi-step reasoning and cross-app information aggregation.
We report Task Goal Completion (TGC) and Scenario Goal Completion (SGC). TGC measures success on an individual task, while SGC measures whether all tasks within a scenario are completed successfully, reflecting broader consistency across related subtasks.

\paragraph{BFCL.}
BFCL (Berkeley Function Calling Leaderboard) evaluates function/tool calling ability, including multi-turn, parallel, and nested tool-use scenarios.
We use the BFCL v3 Multi-turn subset for evaluation and we evaluate models using multi-turn function calling accuracy. A test case is considered successful only if the model selects the correct function and generates semantically and syntactically valid arguments at every turn of the interaction. Any error at an intermediate step results in failure of the entire instance. This metric therefore provides a strict measure of long-horizon tool-use consistency, reflecting the model’s ability to maintain correct function semantics and parameter grounding across multi-step interactions.

\paragraph{WebShop.}
WebShop~\cite{yao2022webshop} is an interactive environment that simulates an e-commerce shopping scenario.
An agent interacts with the environment through two actions, \texttt{search[query]} and \texttt{click[element]}, to complete natural-language shopping requests via product search, attribute filtering, and purchase decisions.
We evaluate performance using the attribute-matching score between the final selected product and the user request.

\paragraph{ALFWorld.}
ALFWorld~\cite{shridhar2020alfworld} is a text-only environment derived from household embodied tasks in ALFRED.
It requires an agent to solve long-horizon tasks in partially observable indoor environments through textual actions for navigation, container operations, and object manipulation.
The task set includes pick-and-place, examination, cleaning, heating, cooling, and multi-object placement scenarios.
We report success rate, where an episode is counted as successful only when the full goal is completed.

\subsection{Implementation Details}
\label{sec:appendix:setup}

\begin{table}[h]
\centering
\setlength{\tabcolsep}{6pt}
\begin{tabular}{@{}ll@{}}
\toprule
\textbf{Parameter} & \textbf{Value} \\
\midrule
Learning rate & $1\mathrm{e}{-6}$ \\
Group size ($n$) & 8 \\
Training batch size & 32 \\
Optimizer & AdamW \\ 
Clip ratio low & 0.20 \\
Clip ratio high & 0.28 \\
KL coefficient & $1\mathrm{e}{-3}$ \\
Rollout temperature & 0.9 \\
Evaluation temperature & 0 \\
Max response length & 4096 \\
Reward signal & success $=1$, failure $=0$ \\
Max interaction step & 30 \\
\bottomrule
\end{tabular}
\caption{Hyperparameters for RL training.}
\label{tab:appendix:RL hparams}
\end{table}

We use the VeRL framework to train the agent with GRPO. The detailed hyperparameters are summarized in Table~\ref{tab:appendix:RL hparams}.
For Qwen2.5-7B-Instruct and Qwen3-4B-Instruct, training is conducted on a single machine equipped with 8$\times$ NVIDIA H20 GPUs (Tensor Parallel = 1), while Qwen3-30B-A3B-Instruct is trained across two machines with 8$\times$ H20 GPUs each (Tensor Parallel = 2).
During training, each interaction episode is capped at 30 environment steps for AppWorld and BFCL, and 15 steps for WebShop and ALFWorld. Exceeding these limits is treated as task failure.
Unless otherwise specified, we initialize the synthetic task set with 100 tasks, train for a total of 120 steps, and regenerate feedback data every 10 training steps. We use Qwen3-Max~\cite{qwen3max} as the exploration LLM.

We compare against closed-source LLMs (Claude Sonnet 4.5~\cite{anthropic2025sonnet45}, GPT-4~\cite{achiam2023gpt}, and Gemini-2.5-Flash~\cite{comanici2025gemini}) and open-source LLMs (DeepSeek-V3.2~\cite{liu2025deepseek}, GPT-OSS-20B~\cite{agarwal2025gpt}, LLaMA-3.3-70B~\cite{grattafiori2024llama}, and Gemma-3-27B~\cite{team2025gemma}). We also report results for backbone models (Qwen2.5-7B-Instruct~\cite{team2024qwen2}, Qwen3-4B-Instruct~\cite{yang2025qwen3}, and Qwen3-30B-A3B-Instruct~\cite{yang2025qwen3}) with and without CoEvolve.

\subsection{Additional Experiments and Analyses}
\label{sec:appendix:extra}

\paragraph{Comparison with adaptive data-generation methods on diverse environments.}
We further evaluate whether the gains of CoEvolve transfer beyond API/function-calling tasks. To this end, we compare CoEvolve with zero-shot, GRPO, and adaptive data-generation baselines, including Reflexion~\cite{shinn2023reflexion} and ReST~\cite{gulcehre2023reinforced}, on ALFWorld~\cite{shridhar2020alfworld}, BFCL, AppWorld, and WebShop~\cite{yao2022webshop} under the same Qwen3-4B-Instruct backbone.

\begin{table*}[t]
\centering
\setlength{\tabcolsep}{7pt}
\begin{tabular}{lcccc}
\toprule
\textbf{Method} & \textbf{ALFWorld} & \textbf{BFCL} & \textbf{AppWorld} & \textbf{WebShop} \\
\midrule
Zero-shot & 30.00 & 26.50 & 16.67 & 5.00 \\
Reflexion & 35.00 & 31.00 & 18.45 & 10.50 \\
Curriculum Learning & 78.57 & 54.50 & 29.17 & 78.60 \\
ReST & 77.94 & 52.00 & 32.74 & 77.20 \\
GRPO & 77.85 & 58.00 & 28.57 & 75.00 \\
CoEvolve & \textbf{82.86} & \textbf{63.00} & \textbf{35.71} & \textbf{80.60} \\
\bottomrule
\end{tabular}
\caption{Comparison with adaptive data-generation methods on four interactive environments using Qwen3-4B-Instruct.}
\label{tab:appendix:adaptive_baselines}
\end{table*}

Table~\ref{tab:appendix:adaptive_baselines} shows that CoEvolve transfers beyond function-calling environments. Under the same Qwen3-4B-Instruct backbone, it consistently outperforms all baselines across ALFWorld, BFCL, AppWorld, and WebShop. In particular, CoEvolve beats Curriculum Learning and ReST~\cite{gulcehre2023reinforced} on ALFWorld and WebShop, and surpasses Reflexion~\cite{shinn2023reflexion} and Curriculum Learning on BFCL and AppWorld. These results suggest that the gains of CoEvolve extend to broader interactive settings such as household decision-making and web navigation.

\paragraph{Task Validation for Abstracted Tasks.}
Table~\ref{tab:appendix:no_validation} examines the role of task validation during abstracted task generation on BFCL (Multi-turn Base) and AppWorld (TestN). In this ablation, we remove the validation step that filters abstracted tasks through environment execution, while keeping all other components unchanged.

Removing task validation leads to a clear and consistent performance degradation across both benchmarks. On BFCL, the score drops from 63.00 to 58.50, while on AppWorld the performance decreases more sharply from 35.71 to 27.38. These results indicate that without validation, a substantial portion of synthesized tasks are either noisy or misaligned with the environment dynamics, which in turn degrades downstream training.

This ablation highlights the importance of validation as a critical component of the feedback-driven data evolution process. By grounding abstracted tasks in actual environment execution, validation ensures that newly added data reflects executable and informative interactions rather than spurious abstractions. As a result, task validation plays a key role in maintaining the quality of the evolving training distribution and enabling effective agent–data co-evolution.

\begin{table}[t]
\centering
\setlength{\tabcolsep}{8pt}
\begin{tabular}{lcc}
\toprule
\textbf{Setting} & \textbf{BFCL} & \textbf{AppWorld} \\
\midrule
Default & \textbf{63.00} & \textbf{35.71} \\
 + w/o validation & 58.50 & 27.38 \\
\bottomrule
\end{tabular}
\caption{Impact of removing validation for abstracted tasks, using Qwen3-4B-Instruct. Metrics are on BFCL Multi-turn Base and AppWorld Test-Normal.}
\label{tab:appendix:no_validation}
\end{table}

\begin{figure*}[t]
    \includegraphics[width=0.95\textwidth]{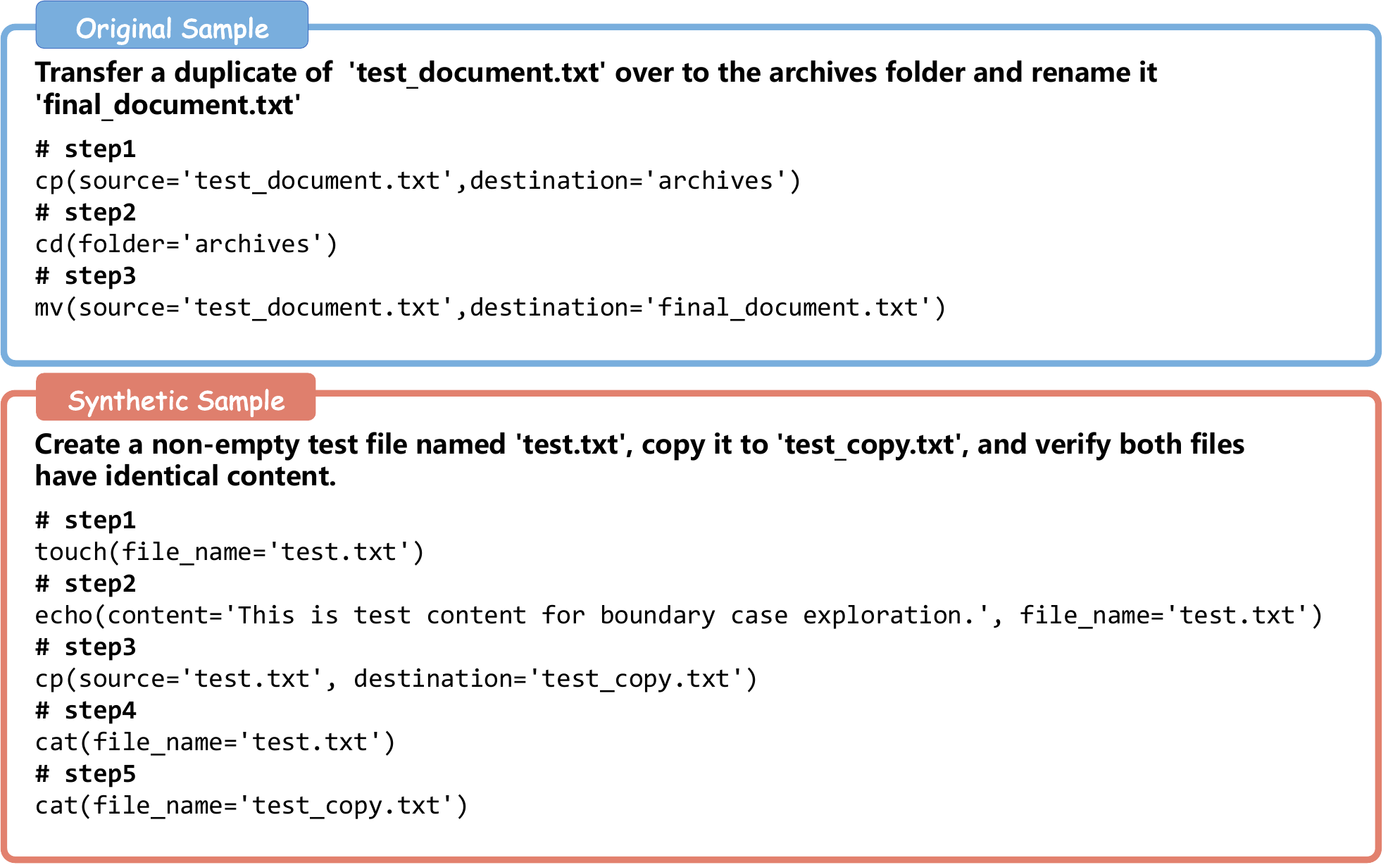}
  \caption{BFCL-V3 cases. Simple File Copy/Rename vs. Constraint-Based Copy with Content Verification}
  \label{fig:sample_a}
\end{figure*}

\paragraph{Controlled study of exploration model quality and feedback.}
Table~\ref{tab:appendix:explore_model} controls the external exploration model used for both initial synthesis and re-exploration on BFCL, while keeping the training model fixed to Qwen3-4B-Instruct. Each row uses the same external model for the ``Synthesis Only'' baseline and the full CoEvolve variant, so the gap isolates the contribution of feedback-guided data evolution rather than model substitution alone.

\begin{table}[t]
\centering
\small
\setlength{\tabcolsep}{5pt}
\begin{tabular}{lcc}
\toprule
\textbf{Exploration Model} & \textbf{Synthesis Only} & \textbf{CoEvolve} \\
\midrule
Qwen3-4B & 53.00 & 56.50 \\
Qwen-Plus & 54.00 & 59.50 \\
Qwen-Max & \textbf{58.00} & \textbf{63.00} \\
\bottomrule
\end{tabular}
\caption{Exploration-model study on BFCL under matched external models.}
\label{tab:appendix:explore_model}
\end{table}

The results support two conclusions. First, stronger exploration models raise the attainable ceiling (Qwen3-4B $<$ Qwen-Plus $<$ Qwen-Max), which is expected and consistent with intuition. Second, under the same exploration model, adding feedback improves over the corresponding ``Only Synthesis'' baseline, showing that CoEvolve benefits from framework design beyond simply swapping in a stronger external model.

\paragraph{Similarity-controlled task synthesis.}
To better understand the relationship between task relevance and final performance, we group synthesized BFCL tasks by their maximum similarity to validation tasks into low, medium, high, and mixed settings. The mean similarity of the low, medium, and high bins is 38.43\%, 53.28\%, and 64.73\%, respectively.

\begin{table}[t]
\centering
\setlength{\tabcolsep}{8pt}
\begin{tabular}{lc}
\toprule
\textbf{Setting} & \textbf{BFCL} \\
\midrule
High-similarity synthesis & 56.50 \\
Medium-similarity synthesis & 59.00 \\
Low-similarity synthesis & 56.50 \\
Mixed synthesis & \textbf{63.00} \\
\bottomrule
\end{tabular}
\caption{Similarity-controlled synthesis on BFCL using Qwen3-4B-Instruct.}
\label{tab:appendix:similarity_bins}
\end{table}

As shown in Table~\ref{tab:appendix:similarity_bins}, these results provide two insights. First, performance is non-monotonic across single similarity bins, suggesting that similarity alone does not determine final performance. Second, the mixed setting performs best, indicating that balancing relevance and diversity across similarity levels is more effective than concentrating synthesis on a single range.

\paragraph{Extended hyperparameter range.}
We further evaluate hyperparameter values beyond the range considered in the main paper to test robustness outside the budgeted setting in Table~\ref{tab:hyperparameter_analysis}. CoEvolve remains reasonably robust beyond the reported range, while more extreme settings mainly introduce a trade-off between data quality and update cadence rather than changing the overall conclusion.

\begin{table}[t]
\centering
\setlength{\tabcolsep}{8pt}
\begin{tabular}{lc}
\toprule
\textbf{Setting} & \textbf{BFCL} \\
\midrule
$N=25$ & 52.00 \\
$N=500$ & 58.00 \\
$F=2$ & 62.00 \\
$F=40$ & 58.00 \\
\bottomrule
\end{tabular}
\caption{Extended hyperparameter sensitivity on BFCL with Qwen3-4B-Instruct.}
\label{tab:appendix:hyper_extreme}
\end{table}

\subsection{Analysis of Interaction Turns.}

\begin{figure}[t]
  \includegraphics[width=\columnwidth]{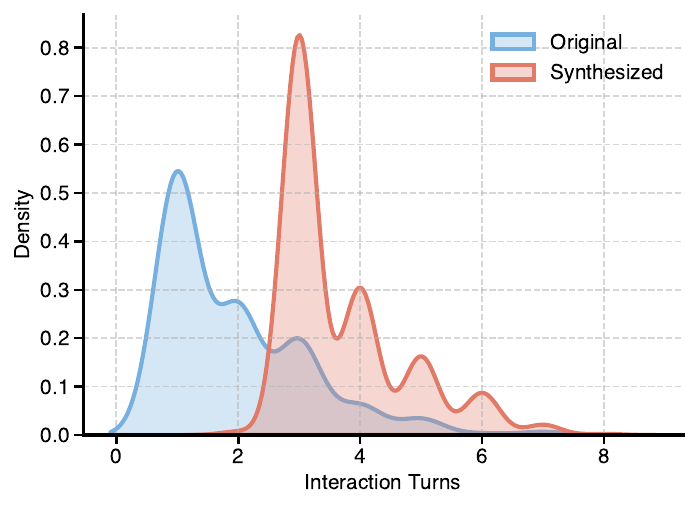}
  \caption{Distribution of interaction turns in BFCL-V3 for original versus synthesized tasks. }
  \label{fig:step_distribution}
\end{figure}

Figure~\ref{fig:step_distribution} compares the distribution of interaction turns between original and synthesized trajectories on BFCL. Relative to the original data, synthesized tasks exhibit a noticeable shift toward higher step counts and a heavier tail, indicating that they more frequently involve longer interaction sequences and multi-step dependencies. This distributional difference suggests that feedback-driven task synthesis tends to generate structurally more complex interaction scenarios, rather than concentrating on the short-horizon tasks that dominate the original dataset. In contrast, the original data shows a more concentrated distribution over interaction length, covering a narrower range of step counts. By introducing tasks with longer interaction sequences, CoEvolve expands the coverage of the training data distribution at the level of interaction structure. Overall, these observations indicate that feedback-driven data evolution can alter the distribution of interaction lengths in a systematic manner. This shift is consistent with the design goal of CoEvolve, which aims to complement static datasets by dynamically discovering underrepresented interaction patterns, without making claims beyond the data distribution itself.

\subsection{Diversity and Relevance Analysis.}
We provide the diversity and relevance metrics used to evaluate the quality of the generated tasks.
\paragraph{Diversity.} Diversity is measured using Self-Redundancy@k (SR@k). Specifically, based on sentence embeddings of the synthesized task intents, we compute, for each task, the average cosine similarity to its $k$ nearest neighbors. Lower SR@k indicates less redundancy among tasks and thus higher diversity. The SR@k metric is calculated as follows:
\begin{equation}
    \mathrm{SR@}k=\frac{1}{|Y|}\sum_{i}\frac{1}{k}\sum_{j\in \mathrm{kNN}_Y(i)}\langle y_i,y_j\rangle
\end{equation}

\paragraph{Relevance.} We measure relevance using the Relative Energy Distance ($\mathrm{ED}_{\mathrm{rel}}$), which quantifies the distributional discrepancy between the target (ground-truth) task-intent distribution (e.g., human-annotated intents or a predefined target distribution) and the generated task intents. Lower $\mathrm{ED}_{\mathrm{rel}}$ indicates that the generated tasks better match the desired/true task distribution. The relative energy distance is:
\begin{equation}
\label{eq:ed_def}
\mathrm{ED}_{\mathrm{rel}}=\frac{\mathrm{ED}(X,Y)}{\mathbb{E}_{i\neq i'}\|x_i-x_{i'}\|_2}
\end{equation}

\begin{equation}
\begin{split}
\mathrm{ED}(X, Y) = 
& \frac{2}{|X||Y|} \sum_{i=1}^{|X|} \sum_{j=1}^{|Y|} \|x_i - y_j\|_2 \\
& - \frac{1}{|X|^2} \sum_{i=1}^{|X|} \sum_{i'=1}^{|X|} \|x_i - x_{i'}\|_2 \\
& - \frac{1}{|Y|^2} \sum_{j=1}^{|Y|} \sum_{j'=1}^{|Y|} \|y_j - y_{j'}\|_2.
\end{split}
\end{equation}

\begin{table}[h!]
\centering
\begin{tabular}{@{}ccc@{}}
\toprule
\textbf{Step} & \textbf{SR (\%)} & \textbf{$\mathbf{ED}_{\mathbf{rel}}$ (\%)} \\ 
\midrule
10  & 22.57 & 0.92 \\
20  & 28.61 & 3.11 \\
30  & 16.16 & 0.76 \\
40  & 19.11 & 0.24 \\
50  & 26.14 & 1.65 \\
60  & 21.24 & 0.79 \\
70  & 17.91 & 0.37 \\
80  & 19.82 & 0.50 \\
90  & 22.78 & 1.20 \\
100 & 14.76 & -0.11 \\
110 & 22.22 & 0.46 \\
120 & 25.92 & 1.49 \\
\midrule
Mean & 21.44 & 0.95 \\
Variance & 17.30 & 0.73 \\
\bottomrule
\end{tabular}
\caption{Trend of SR and $\mathrm{ED}_{\mathrm{rel}}$ Across Steps}
\label{tab:appendix:data_relevrance_diversity}
\end{table}

\begin{figure*}[t]
    \includegraphics[width=0.95\textwidth]{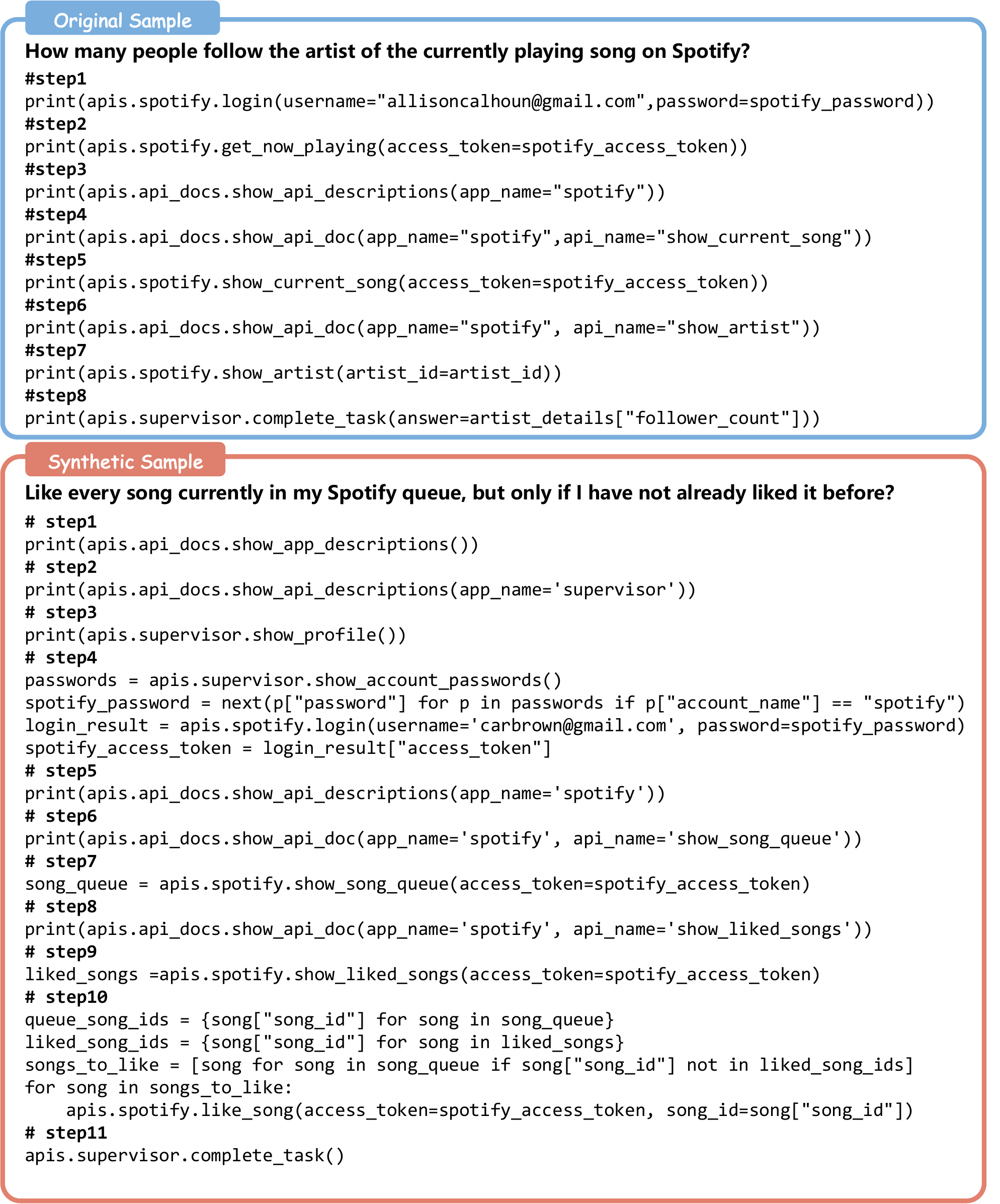}
  \caption{AppWorld cases. Now-Playing Artist Followers Lookup vs. Conditional “Like Queue” with Dedup Filtering.}
  \label{fig:sample_b}
\end{figure*}

Table~\ref{tab:appendix:data_relevrance_diversity} presents the trends of SR (Self-Redundancy) and $\mathrm{ED}_{\mathrm{rel}}$ (Relative Energy Distance) across training steps. The mean SR is 21.44\%, indicating a moderate level of redundancy among synthesized task intents, while the relatively high variance (17.30\%) highlights significant fluctuations in redundancy across different steps. Similarly, the mean $\mathrm{ED}_{\mathrm{rel}}$ is 0.95\%, reflecting a stable alignment between the generated and target task distributions, with a much lower variance (0.73\%) compared to SR. Such observations suggest distinct behaviors of redundancy and distribution alignment within the synthesis process.

\subsection{Synthetic Sample Analysis.}
We show representative synthetic and validation examples from AppWorld (Fig.~\ref{fig:sample_b}) and BFCL-V3 (Fig.~\ref{fig:sample_a}). On BFCL, the original sample reflects a typical short-horizon tool-usage pattern (3 steps: copy → cd → rename) where success is mostly determined by the final state, while the synthetic sample is a longer-horizon BFCL-style task (5 steps) that explicitly requires state construction and correctness verification (create a non-empty file, copy it, then inspect both files to confirm identical content). This increases the number of interaction rounds, tightens step dependencies, and introduces explicit constraints and validation, thereby better stressing multi-round task planning and correctness checking. On AppWorld, the original sample is essentially a single-query information retrieval task wrapped in tool calls: log into Spotify, fetch the currently playing track, look up the artist, and return the artist’s follower count—despite multiple API-doc lookups, the logic is mostly linear and the success criterion is one scalar value. In contrast, the synthetic sample is a multi-step state-changing workflow with conditional control: it must authenticate via the supervisor/password flow, fetch the current queue, fetch the user’s liked songs, compute a set difference to identify only unliked tracks, and then iterate to like each remaining song. This increases interaction rounds (about 8 vs. 11), introduces cross-endpoint state alignment (queue vs. liked library), adds non-trivial intermediate computation (ID extraction and filtering), and carries higher risk (avoiding duplicate likes), highlighting the greater compositional complexity and longer-horizon execution that synthetic AppWorld tasks are designed to stress.

Across both BFCL and AppWorld, the original samples are mostly linear, short-horizon tasks with simple end-state or single-answer goals, while the synthetic samples require more rounds, stronger cross-step dependencies, intermediate reasoning (e.g., filtering/set-difference), and explicit correctness constraints/verification—therefore better reflecting higher compositional complexity and longer-horizon tool-use.

\subsection{Prompts Used in the Feedback Loop.}
\label{sec:appendix:prompts}
\begin{table}[t]
\centering
\small
\setlength{\tabcolsep}{4pt}
\begin{tabular}{@{}>{\raggedright\arraybackslash}p{0.58\columnwidth}>{\raggedright\arraybackslash}p{0.20\columnwidth}@{}}
\toprule
\textbf{Module} & \textbf{Prompt Figure} \\
\midrule
Exploration system/user prompts & Fig.~\ref{fig:template_exploration_all} \\
Signal-specific exploration guidance & Fig.~\ref{fig:prompt_templates_all_vertical} \\
Signal-conditioned context summarization & Fig.~\ref{fig:template_signal_summary} \\
Task abstraction & Fig.~\ref{fig:template_task_abstraction} \\
Task validation & Fig.~\ref{fig:template_task_validation} \\
\bottomrule
\end{tabular}
\caption{Prompt-to-module mapping for feedback loop.}
\label{tab:appendix:prompt_mapping}
\end{table}
\begin{figure}[h]
    \centering

    \begin{templatebox}{(a) System Prompt Template for Exploration} 
    You are exploring based on specific guidance to help improve an AI agent's capabilities.

    ## Your Task:
    1. Observe the current environment state and identify available actions
    2. Analyze available actions and determine which ones will help with the exploration goal
    3. Select a relevant action and execute it in the required format
    4. Focus on thorough exploration of the targeted area

    # Action Format:
    {action_format}

    ## Instructions:
    - Choose only one action at a time
    - Carefully read the environment description and task instructions
    - Ensure that the action is in the correct format
    - Do not use undefined actions
    - Always include a valid action and action tags in your reply
    - First enter your reason, then enter your action
    \end{templatebox}

    \vspace{5.5em}

    \begin{templatebox}{(b) User Prompt Template for Exploration} 
    {exploration_guidance} 

    ## Environment Description:
    {initial_obs}
    
    ## Recent History:
    {history_text}
    
    Please select an appropriate action based on the exploration goal and current state.
    \end{templatebox}

    \caption{Prompt templates for exploration: (a) system-side prompt and (b) user-side prompt.}
    \label{fig:template_exploration_all}
\end{figure}

\begin{figure}[t]
    \centering

    \begin{templatebox}{(a) Template for Forgetting Signal Guidance} 
Exploration Goal: Reinforce Forgotten Skills
The agent previously succeeded but now FAILS on this type of task.
Your exploration should:
1. Practice the exact operations from the context below
2. Create variations with different parameters
3. Connect this skill to related operations
4. Build up from simple to complex usage
Context of forgetting:
{context}
Focus on thorough practice of these specific operations.
    \end{templatebox}


    \begin{templatebox}{(b) Template for Rare Event Signal Guidance} 
Exploration Goal: Explore Rare Scenarios
The agent encountered a RARE scenario that needs more exposure.
Your exploration should:
1. Explore variations of the scenario below
2. Try different parameter combinations
3. Test edge cases and boundary conditions
4. Collect diverse examples of this rare pattern
Context of rare event:
{context}
Try to discover and document various forms of this scenario.
    \end{templatebox}


    \begin{templatebox}{(c) Template for Boundary Case Signal} 
Exploration Goal: Explore Boundary Cases
The agent's performance is BORDERLINE (near success/failure threshold).
Your exploration should:
1. Explore boundary conditions for these operations
2. Try similar tasks with slight parameter variations
3. Focus on distinguishing factors between success and failure
4. Collect examples at various difficulty levels
Context of boundary case:
{context}
Focus on understanding what makes the difference between success and failure.
\end{templatebox}
\caption{Prompt templates for three types of exploration signals:
(a) forgetting, (b) rare event, and (c) boundary case.}
    \label{fig:prompt_templates_all_vertical}
\end{figure}

\begin{figure*}[h]
    \centering
    \begin{templatebox}{Prompt Template for Signal-Conditioned Context Summarization}
You are an expert at analyzing trajectory-level failure and behavioral instability for an LLM agent.

Analyze the following feedback signal and trajectory evidence:

Feedback Signal:
- Signal: {signal}

Trajectory Evidence:
{trajectory_context}

Your goal is to extract a structured exploration context that captures:
1) A concise recap of this trajectory
2) Why failure or instability happened
3) Which patterns or behaviors should be re-explored
4) What mistakes should be explicitly avoided

Return a JSON object with this schema:
{
  "summary": "concise recap of the current trajectory evidence (task, key actions, and feedback outcome)",
  "failure_cause": "1-3 sentence root cause of failure or instability",
  "instability_pattern": "1-2 sentence pattern summary",
  "focus_pattern": ["pattern or behavior to focus on", "..."],
  "exploration_objectives": ["concrete exploration objective", "..."],
  "do_not_repeat": ["common mistake to avoid", "..."]
}

Rules:
- Ground every statement in the provided evidence.
- Keep the output concise and actionable.
- First produce `summary` from the trajectory evidence, then derive the other fields from that summary.
    \end{templatebox}
    \caption{Prompt template for signal-conditioned trajectory summarization.}
    \label{fig:template_signal_summary}
\end{figure*}

\begin{figure*}[h]
    \centering
    \begin{templatebox}{Prompt Template for Task Abstraction} 
    You are a *Task Abstraction Expert*. Your specialty is to inspect an agent's
    interaction history and distill concrete, goal-oriented tasks from it.

    ========================  YOUR JOB  ========================
    1. Inspect the interaction tuples (history, action, observation).
    2. Identify the specific goal or task the agent is attempting to achieve.
    3. Abstract each goal into a clear, concise **task description**, a **query**
        (suitable for search or training), and the **minimal action sequence**
        that successfully completes the task.

    =====================  ABSTRACTION RULES  ==================
    - Focus on clear, goal-directed behaviour; ignore purely random exploration.
    - Please include as many steps as possible in ActionSequence.
    - Group similar behaviour patterns into the same task.
    - Every task must have **at least one** action sequence that was executed successfully.
    - Each task needs an explicit completion criterion.
    - All actions listed in an action sequence must be valid and directly executable.
    - Ensure all actions are combined into a minimum sequence from initial state to completion.
    - The ActionSequence should have at least 3 steps.

    ========================  OUTPUT FORMAT  ===================
    For every task you identify, output exactly one block in the form below:
    {output_format}
    \end{templatebox}
    \caption{Prompt template for task abstraction.}
    \label{fig:template_task_abstraction}
\end{figure*}

\begin{figure*}[h]
    \centering
    \begin{templatebox}{Prompt Template for Task Validation} 
    You are a strict task evaluation expert. Your goal is to determine whether the following multi-step agent trajectory successfully completed the assigned task.

    # Task Details
    - Task Description: {task_description}
    - Query: {query}
    - Expected Outcome (API Call or Result): {ground_truth}
    - Action Modality: {modality_hint}
    
    # Execution Summary
    - Trajectory Summary:
    {trajectory_summary}
    
    - Final Observation: {final_observation}
    
    # Evaluation Instructions
    
    Carefully analyze the trajectory to determine if the task was truly completed. Specifically, consider the following aspects:
    
    1. **API Matching**: Did the agent correctly call the required APIs according to the task requirements?
    2. **Parameter Usage**: Were the parameters used in API calls correct and sufficient?
    3. **Logical Flow**: Was the sequence of steps logical without unreasonable skips?
    4. **Final Result**: Did the final state achieve the expected outcome, reasonably solve the task, obtain all necessary information, and complete the task objectives?
    5. **Failed or Skipped Steps**: Were there any critical errors, skipped steps, or invalid code that prevented the task from being actually executed?
    
    # Format Your Response Strictly As:
    
    Success: [true/false]
    Reason: [Concise and specific explanation, referring to the above criteria.]
    
    Note: Ignore all Connection timeout or No valid action, because it is very likely that it is the former. Do NOT mark the task as successful if the correct API was never called, the parameters were incorrect, or the result was not achieved, even if the intent seemed right.    \end{templatebox}
    \caption{Prompt template for task validation.}
    \label{fig:template_task_validation}
\end{figure*}

We briefly describe the prompt templates used throughout the exploration and data evolution stages. These prompts define how the exploration model is instructed to generate candidate trajectories, interpret different feedback signals, and transform raw interaction traces into reusable training tasks.
Table~\ref{tab:appendix:prompt_mapping} maps each critical LLM call to its corresponding template.

Specifically, Fig.~\ref{fig:template_exploration_all} shows the general prompt templates used for exploration, including the system-side prompt that specifies the role and constraints of the exploration model, and the user-side prompt that provides task context and feedback information. These prompts establish the basic interaction protocol for task proposal.

To specialize exploration toward different failure modes, we further design signal-conditioned prompt templates for three types of training-time feedback signals: forgetting, rare events, and boundary cases, as illustrated in Fig.~\ref{fig:prompt_templates_all_vertical}. Each template explicitly conditions the exploration process on the corresponding signal, encouraging the model to generate tasks that target the agent’s observed weaknesses.

Fig.~\ref{fig:template_signal_summary} shows the prompt used for this signal-conditioned summarization step. Given the signal type and the full trajectory evidence, it extracts a concise recap of the failure case, identifies the likely failure cause or instability pattern, and produces structured fields such as focus patterns, exploration objectives, and ``do-not-repeat'' constraints. This intermediate representation serves as the bridge between low-level rollout traces and the downstream exploration prompts, ensuring that subsequent exploration is grounded in concrete behavioral evidence rather than loosely conditioned on the signal name alone.

After candidate tasks are proposed, task validation and abstraction are handled by dedicated prompt templates. Fig.~\ref{fig:template_task_validation} presents the prompt used to verify task executability through environment interaction, ensuring that only valid tasks are retained. Fig.~\ref{fig:template_task_abstraction} shows the abstraction prompt, which converts validated interaction traces into concise and reusable task specifications suitable for training.

\subsection{Use of Large Language Models.}
During manuscript preparation, we use large language models (LLMs) to (i) improve grammar and spelling without altering the intended scientific content, and (ii) provide lightweight coding assistance (e.g., scripts and formatting help). All reported numerical results, analyses, and claims are produced by the authors. The authors design the methods, conduct the experiments, and verify the findings.

\end{document}